\documentclass[lettersize,journal]{IEEEtran}

\usepackage{amsmath,amsfonts}
\usepackage{algorithm}
\usepackage{algorithmic}
\usepackage{array}
\usepackage[caption=false,font=normalsize,labelfont=sf,textfont=sf]{subfig}
\usepackage{textcomp}
\usepackage{stfloats}
\usepackage{url}
\usepackage{verbatim}
\usepackage{graphicx}
\usepackage{cite}
\usepackage{bibunits}
\usepackage{balance}
\usepackage{placeins}
\usepackage[table]{xcolor}
\usepackage{booktabs}
\usepackage{adjustbox}
\usepackage{array}

\usepackage{multirow}
\usepackage{makecell}
\usepackage{siunitx}
\usepackage{adjustbox}
\usepackage{pifont}
\newcommand{\cmark}{\ding{51}}

\usepackage{tcolorbox}
\usepackage[hidelinks]{hyperref}

\usepackage{xcolor}

\hypersetup{
    pdfauthor={Jia Li, Yu Zhang, et al.},
    pdftitle={Bidirectional Learning of Facial Action Units and  Expressions via Structured Semantic Mapping across Heterogeneous Datasets},
    pdfsubject={Affective Computing},
    pdfkeywords={affective computing, facial action units, dynamic facial expression, semantic mapping},
    colorlinks=true,
    citecolor=blue,
    linkcolor=blue,
    urlcolor=blue
}

\begin{document}


\title{Bidirectional Learning of Facial Action Units and  Expressions via Structured Semantic Mapping across Heterogeneous Datasets}

\author{Jia~Li, Yu~Zhang, Yin~Chen, Zhenzhen~Hu, Yong~Li, Richang~Hong,~\IEEEmembership{Member,~IEEE,} Shiguang~Shan,~\IEEEmembership{Fellow,~IEEE,} and Meng~Wang,~\IEEEmembership{Fellow,~IEEE}%
	\thanks{
    
        This work was supported in part by the New Cornerstone Science Foundation through the XPLORER PRIZE, and also in part by the National Natural Science Foundation of China (No. U2336213, 62576099)

		Jia~Li, Yu~Zhang, Yin~Chen, Zhenzhen~Hu, Richang~Hong and Meng~Wang are with the School of Computer
		Science and Information Engineering, Hefei University of Technology, Hefei
		230601, China (e-mail: jiali@hfut.edu.cn; yuz@mail.hfut.edu.cn; chenyin@mail.hfut.edu.cn; huzhen.ice@gmail.com; hongrc.hfut@gmail.com; eric.mengwang@gmail.com).
		
		Yong Li is with the School of Computer Science and Engineering, and the
        Key Laboratory of New Generation Artificial Intelligence Technology and Its Interdisciplinary Applications, Southeast University, Nanjing 210096, China (e-mail: mysee1989@gmail.com).
		
		Shiguang~Shan is with the Key Laboratory of Intelligent Information Processing,
		Institute of Computing Technology, Chinese Academy of Sciences, Beijing 100190,
		China, and also with the University of Chinese Academy of Sciences, Beijing 100049, China (e-mail: sgshan@ict.ac.cn).
		
		(Corresponding author: Zhenzhen~Hu.)
	}
}


\maketitle

\begin{abstract}

Facial action unit (AU) detection and facial expression (FE) recognition  can be jointly viewed as affective facial behavior tasks, representing fine-grained muscular activations and coarse-grained holistic affective states, respectively. Despite their inherent semantic correlation, existing studies  predominantly  focus on knowledge transfer from AUs to FEs, while  bidirectional learning remains insufficiently explored. In practice, this challenge is further compounded by heterogeneous data conditions, where AU and FE datasets differ in annotation paradigms (frame-level vs.\ clip-level), label granularity, and data availability and diversity, hindering effective joint learning.
To address these issues, we propose a Structured Semantic Mapping (SSM) framework for bidirectional AU--FE learning under different data domains and  heterogeneous supervision. SSM consists of three key components: (1) a shared visual backbone that learns unified facial representations from dynamic AU and FE videos; (2) semantic mediation via a Textual Semantic Prototype (TSP) module, which constructs structured semantic prototypes from fixed textual descriptions with learnable context prompts for supervision and cross-task alignment in a shared semantic space; and (3) a Dynamic Prior Mapping (DPM) module that incorporates FACS-derived prior knowledge and learns data-adaptive bidirectional association matrices in the textual semantic space for explicit knowledge transfer.
Extensive experiments on popular AU detection and FE recognition benchmarks show that SSM consistently outperforms its single-task and multi-task baselines and achieves competitive performance against task-specific methods. The FE→AU results further show that holistic expression semantics provides useful supervision for fine-grained AU learning across heterogeneous datasets.

\end{abstract}


\section{Introduction}



\begin{figure}[t]  
    \centering
    \includegraphics[width=\columnwidth]{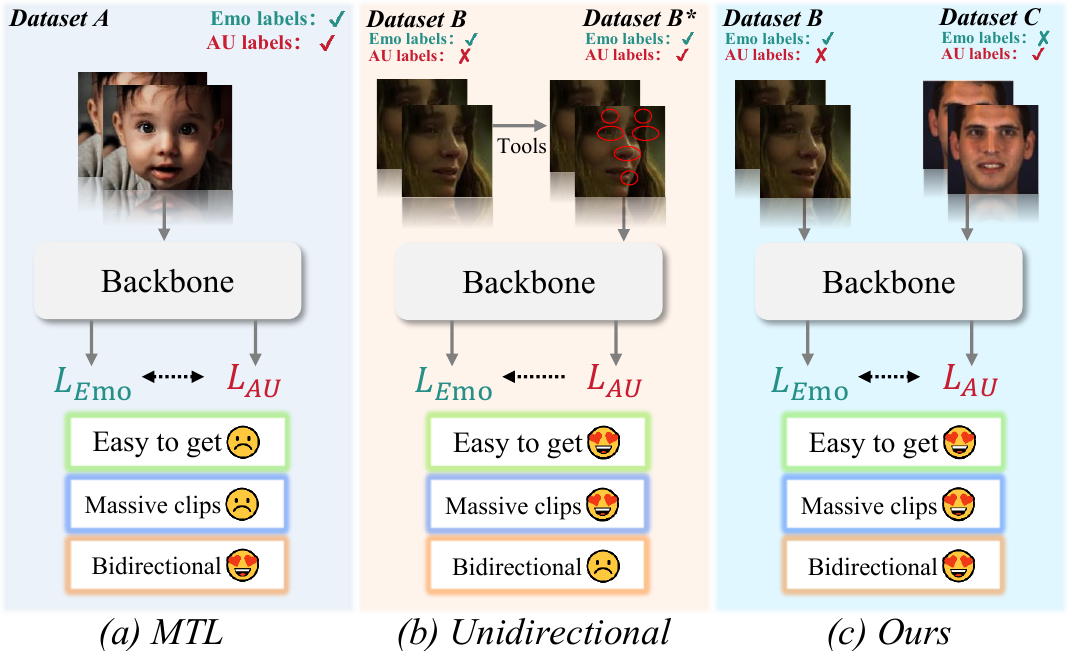}
\caption{Paradigm comparison of joint AU and FE learning. 
(a): conventional multi-task learning (MTL) on homogeneous datasets achieves bidirectional gains but suffers from high annotation cost and limited generalization.
(b): predominant unidirectional transfer (AU$\rightarrow$FE) on heterogeneous datasets, where mismatched annotation paradigms and semantic granularity hinder reverse learning and introduce distribution bias from lab-controlled AU data. 
(c): our method enables adaptive bidirectional learning by explicitly modeling AU$\leftrightarrow$FE relationships via textual mediation in a shared semantic space, improving scalability and generalization under heterogeneous data.}
    \label{intro_image}
\end{figure}


\IEEEPARstart{D}{eeply} understanding human emotions, intentions, and social signals requires comprehensive analysis of affective facial behaviors, which plays a critical role in applications such as human–computer interaction, social robotics, mental health monitoring, and driver safety. From a psychological perspective, facial expressions (FEs) exhibit a certain degree of universality across cultures \cite{ekman1971constants,matsumoto1992more}, while being fundamentally driven by coordinated facial muscle movements, namely, Action Units (AUs). The well-known Facial Action Coding System (FACS) \cite{ekman2002facs} provides anatomically grounded definitions of facial action units
and their associations with prototypical facial expressions. Accordingly, dynamic facial expression recognition (DFER) and AU detection in videos can be jointly viewed as two core affective facial behavior tasks, corresponding to coarse-grained holistic affective states and fine-grained muscular activations, respectively. Their intrinsic semantic correlation suggests a natural potential for complementary modeling.

In recent years, a large number of supervised-learning-based methods have achieved promising performance on AU detection and DFER \cite{yang2018facial,liu2015spontaneous,ruan2020deep,she2021dive,yang2021exploiting,jacob2021facial,chang2022knowledge}. However, existing large-scale datasets often do not overlap in task annotations, modalities, or domains (i.e., only abundant heterogeneous data available), as depicted in Fig. \ref{intro_image}. Consequently, the prevailing research paradigm has mostly focused on unidirectional facilitation, which uses AU features or statistics as auxiliary signals to improve facial expression recognition. However, this paradigm simply treats facial expressions as mechanical combinations of multiple AU activations. For example, Li et al. \cite{li2023compound} have constructed a knowledge matrix from a dataset's statistics and enhanced the expression task via loss injection, yet the obtained prior is static. This matrix depends on a specific data distribution and is thus susceptible to dataset bias.  Additionally, Kollias et al. \cite{kollias2023multi} have improved compound expression recognition by letting the expression branch predict AU distributions to guide the model in learning the association between the two tasks, but its pseudo-label-based learning mainly remains at the level of shallow feature interactions and implicit fusion. With the increasing availability of video-level datasets, it has gradually been recognized that studying AUs or FEs under dynamic settings is more reliable. Notably, AUs not only characterize static muscular configurations but also reflect dynamic variations during expression generation \cite{chen2023enhanced,mao2022aware,tian2001recognizing}. Therefore, jointly studying AU detection and DFER in 
videos and modeling local and global affective semantics together along the spatio-temporal dimension better accords with the natural physiological mechanisms. At present, studies on  AU$\to$FE knowledge transfer under dynamic settings have preliminarily verified that local facial actions can effectively support global expression understanding \cite{liu2025action}. Hence, a critical question remains: \textbf{\textit{Can the established bidirectional relationship
(AU$\leftrightarrow$FE) between the two tasks be effectively exploited
in heterogeneous dynamic videos?}}

On the other hand, some works rely on expensive multi-label datasets whose annotations, modalities, and domains overlap (i.e., homogeneous datasets).  And they directly perform multi-task training within the same data domain to achieve bidirectional facilitation between AU detection and DFER \cite{ma2023unified, zhang2024effective, kim2024advanced}. For instance,
Kollias et al.\cite{ kollias2019expression,kollias2022abaw} have constructed the Aff-Wild2 dataset, simultaneously providing  frame-level expression categories, continuous valence–arousal values, and AU activations. Then, they propose  a multi-task learning framework, demonstrating that jointly learning multiple facial affective tasks on the same data domain can yield reciprocal benefits.
Based on homogeneous datasets such as Aff-Wild2, existing methods directly perform joint AU--FE learning with multi-task supervision and improve both AU detection and facial expression recognition (FER) \cite{zhang2021prior,jin2021mtmsn,jeong2022multi,zhang2022transformer,savchenko2024hsemotion}. However, these methods entail substantial annotation costs and suffer from domain limitations: frame-level multi-label annotation requires professional FACS coders to annotate videos frame by frame, which is costly, time-consuming, and difficult to scale. Meanwhile, the synergistic gains obtained on small-scale homogeneous data 
can not guarantee model's generalization to real-world scenarios. Hence, it is necessary to explore adaptive mutual-promotion mechanisms between the two tasks under heterogeneous data conditions.

Based on this background, and given the availability of large-scale heterogeneous datasets, a more practical question arises: \textbf{\textit{Can these heterogeneous datasets be effectively leveraged to achieve  AU$\leftrightarrow$FE bidirectional learning benefits?}}

Therefore, we first construct a Baseline model, performing multi-task learning on heterogeneous data, and systematically explore whether there exist stable complementary and reciprocal effects between AU detection and DFER.
However, achieving this goal is non-trivial due to inherent heterogeneity across datasets.
First, the two types of datasets differ in collection environments and annotation systems, leading to inconsistent semantic spaces. Second, the correspondence between AUs and FEs is not mechanical, and activation patterns vary significantly across individuals and laboratory-controlled and in-the-wild scenarios  \cite{li2023compound, kollias2023multi}, making static priors difficult to generalize to uncontrolled real-world scenarios. Third, existing multi-task learning based on shared features or joint losses is prone to negative transfer on heterogeneous data \cite{lu2024fedhca2,khan2024heterogeneous,jain2023damex}. Thus, a further question arises: \textbf{\textit{Can cross-task knowledge be mediated in a shared semantic space to mitigate interference from heterogeneous data distributions?}}

To this end, we propose a Structured Semantic Mapping (SSM) framework to enable bidirectional learning between AU detection and DFER over heterogeneous datasets. Built upon the aforementioned multi-task learning Baseline, SSM introduces a shared semantic space based on textual embeddings to mediate cross-task knowledge and mitigate data heterogeneity. Specifically, SSM further employs two key components: a Dynamic Prior Mapping (DPM) module and a Textual Semantic Prototype (TSP) module. DPM, initialized from FACS priors, learns dynamic and bidirectional correspondences between AUs and FEs in the semantic space explicitly, which are continuously updated during training rather than fixed by dataset statistics. TSP constructs structured semantic prototypes for both tasks, where AU prototypes are directly derived from FACS-defined AU descriptions and FE prototypes are composed based on FACS knowledge, enabling unified semantic encoding and alignment.
Unlike prior works that rely on static statistical priors or shallow feature-level interactions \cite{kollias2023multi,liu2025action}, our framework performs semantic-level knowledge mediation to adaptively capture asymmetric and dataset-dependent AU--FE relationships. This design reduces reliance on homogeneous annotations and mitigates interference caused by heterogeneous data distributions. The source code and models are publicly available here\footnote{\url{https://github.com/MSA-LMC/SSM}}.


Our main contributions are summarized as follows:
\begin{itemize}
    \item We conduct a systematic study of bidirectional AU$\leftrightarrow$FE learning under heterogeneous dynamic settings, demonstrating consistent mutual gains and further highlighting the contribution of DFER to AU detection.
    
    \item We propose a Structured Semantic Mapping (SSM) framework for bidirectional AU$\leftrightarrow$FE transfer without paired multi-task annotations. SSM integrates Textual Semantic Prototype (TSP) and Dynamic Prior Mapping (DPM) in a shared semantic space to enable FACS-guided and data-adaptive cross-task knowledge transfer.
    
    \item Extensive experiments on multiple in-the-wild DFER datasets (DFEW \cite{jiang2020dfew}, MAFW \cite{liu2022mafw}, FERV39K \cite{wang2022ferv39k}) and representative in-the-lab AU detection datasets (BP4D \cite{zhang2014bp4d}, DISFA \cite{mavadati2013disfa}) demonstrate consistent improvements on both tasks across different dataset combinations, together with competitive performance against task-specific state-of-the-art methods for DFER and AU detection.
\end{itemize}
    
    




\section{Related Work}

\subsection{Dynamic Facial Expression Recognition}
Dynamic Facial Expression Recognition (DFER) aims to model the spatial-temporal evolution of facial expressions from videos (or frame sequences in other words). It is a fundamental task in facial behavior analysis. Early methods mainly relied on handcrafted features and shallow temporal models. Recent studies increasingly adopt end-to-end deep learning models to jointly capture spatial and temporal dynamics. A mainstream line of research follows the supervised learning paradigm. It combines convolutional neural networks with temporal modeling modules such as LSTM or Transformer to improve recognition performance \cite{ma2023logo,li2023intensity,liu2023expression,wang2025lifting}. For example, Former-DFER \cite{zhao2021former} integrates spatial convolutional features with a temporal Transformer. It demonstrates strong robustness under challenging conditions.

With the emergence of vision--language pretrained models, recent studies have explored the incorporation of cross-modal semantic knowledge into DFER \cite{tao20243,zhang2024clip,chen2024finecliper,liang2026clvsr}. Methods such as CLIPER \cite{li2024cliper}, DFER-CLIP \cite{zhao2023prompting}, and PE-CLIP \cite{saadi2025pe} leverage the text--vision alignment capability of CLIP \cite{radford2021learning} to project expression categories into a shared semantic space. This design improves generalization despite the lack of domain-specific 
pretraining for facial expressions. In parallel, self-supervised and pretraining-based methods have also been investigated \cite{sun2023mae,cheng2025vaemo}. For instance, MAE-DFER learns discriminative temporal representations through masked reconstruction with a local--global interactive Transformer encoder. S2D \cite{chen2024static} and S4D \cite{chen2025static} transfer knowledge from static expression datasets to dynamic scenarios through self-supervised pretraining and task adaptation. Despite these advances, most existing methods still treat AU detection and DFER as isolated tasks.

\subsection{Facial Action Unit Detection}
Facial Action Unit (AU) detection aims to recognize local facial muscle activations. It is a fine-grained task in facial behavior analysis. Recent studies on dynamic AU detection mainly focus on three aspects: enhancing feature representations, modeling dependencies among AUs, and improving robustness and generalization. First, several studies introduce structured priors or generative modeling to enhance feature representations. These methods alleviate the disturbance of pose variation, occlusion, and cross-dataset discrepancies \cite{song2021hybrid,yang2021exploiting,tang2021piap}. 
Second, another important research direction is modeling dependencies among AUs. Graph neural networks (GNNs) are widely employed to capture AU co-occurrence relationships and structural constraints \cite{li2019semantic,luo2022learning,huang2025facial,hu2025causalaffect}.
Third, with the increasing availability of unlabeled data, self-supervised pretraining has been utilized to learn more powerful  AU representations \cite{ning2024representation,ma2024facial}. To further improve robustness in real-world scenarios, uncertainty modeling mechanisms have also been introduced to improve robustness \cite{song2021uncertain}. 

In addition, multimodal and multi-view learning have also been explored to further improve AU detection performance \cite{yang2020adaptive,zhang2021multi,chang2022knowledge,li2023disagreement,zhang2023weakly,li2026hierarchical}. Despite these advances, most existing methods still focus on the AU task itself. They seldom exploit coarse-grained expression semantics to provide complementary supervision.

\subsection{AU and FE Relationship Modeling}
Early studies mainly follow a unidirectional paradigm in which AUs are treated as auxiliary supervision or intermediate representations to facilitate expression recognition. For instance, Kollias et al. \cite{kollias2023multi} guide the expression branch by predicting AU distributions. However, this approach primarily relies on pseudo labels and tends to operate at the level of shallow feature interactions. In contrast, Li et al. \cite{li2023compound} introduce a static AU--expression knowledge matrix derived from dataset statistics, which is inherently sensitive to data distributions and thus may generalize poorly across different dataset domains.


Another line of work explores homogeneous multi-label datasets, where joint AU detection and FE recognition are achieved via multi-task learning \cite{ma2023unified,zhang2024effective,kim2024advanced}. Studies based on the Aff-Wild2 dataset \cite{kollias2019expression,kollias2022abaw} (contains 558 videos in the wild) have shown that joint optimization on frame-level multi-label annotations can facilitate shared representation learning and lead to mutual performance gains. However, such methods typically rely on densely annotated data  \cite{kollias2022abaw}, which require substantial annotation effort and are not always readily available at scale. Moreover, existing datasets are often constrained in terms of data diversity and accessibility, which may limit their applicability to broader real-world scenarios.

Beyond direct multi-task learning on homogeneous data, knowledge-guided joint AU--FE learning has also been explored. Cui et al.~\cite{cui2020knowledge} encode expression--AU dependencies with a Bayesian network, but the resulting knowledge model remains fixed during subsequent joint training and cannot adapt to a specific heterogeneous dataset pair. Kollias et al.~\cite{kollias2024distribution} perform distribution matching and soft co-annotation under limited or non-overlapping annotations, but the task relatedness used in their coupling losses cannot jointly adapt to the current AU and FE datasets.

Therefore, bidirectional AU--FE relationship modeling under heterogeneous dynamic-video supervision requires further study.


\begin{figure*}[ht!]
  \centering
  \includegraphics[width=\textwidth,height=\textheight,keepaspectratio]{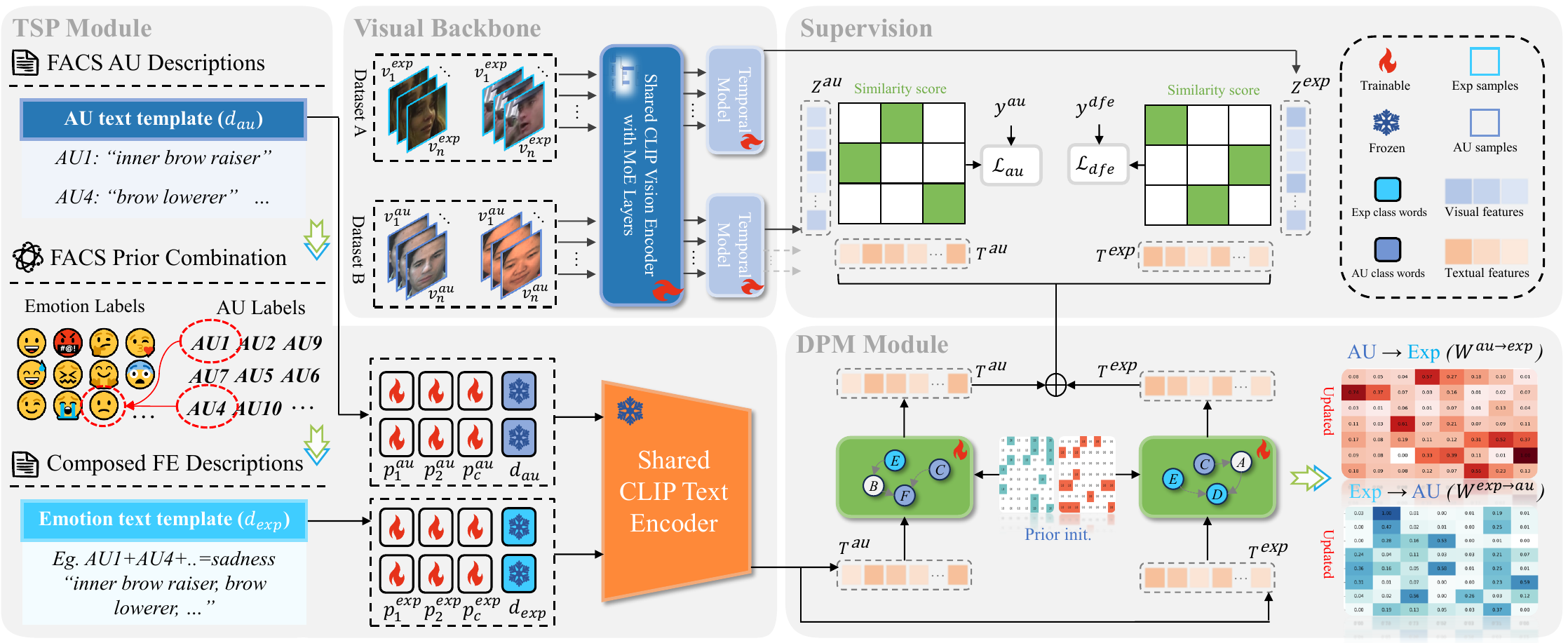}
  \caption{Overview of the Structured Semantic Mapping (SSM) framework. SSM reformulates joint AU detection and DFER in a unified vision--text semantic space. 
The visual branch employs a shared CLIP vision encoder with MoE layers (i.e., our Baseline model) and two task-specific temporal models, producing a clip-level representation for DFER and a temporally enhanced center-frame representation for AU detection from heterogeneous datasets. 
In the textual branch, the Textual Semantic Prototype (TSP) module constructs AU prototypes from FACS-defined AU descriptions and
composes expression prototypes from FACS-derived AU--expression
configurations, which are then encoded by a shared CLIP text encoder. 
Built upon these representations, the Dynamic Prior Mapping (DPM) module performs prior-initialized bidirectional semantic mapping via two learnable association matrices, generating dynamically updated textual representations for cross-task knowledge transfer and contrastive supervision.}
  \label{fig:oview}
\end{figure*}

\section{Method}

From the perspective of unified semantic modeling of AUs and FEs, this paper proposes a cross-task learning framework on heterogeneous data, which aligns the semantics of fine-grained action units and coarse-grained facial expressions without relying on homogeneous multi-label annotations. 
In this section, we first introduce a powerful multi-task Baseline model and the basic concept of CLIP-style prompt learning for classification, and then describe the technical details of the proposed SSM framework, depicted in Fig.~\ref{fig:oview}.

\subsection{Preliminary}

\subsubsection{Baseline Model}
\label{sec:Baseline}

\begin{figure}[t]
    \centering
    \includegraphics[width=\columnwidth]{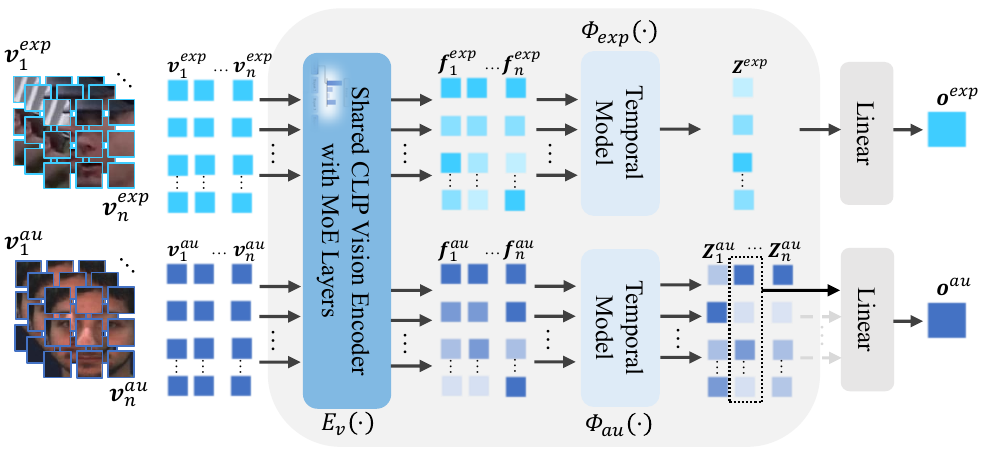}
    \caption{Our Baseline model for  learning dynamic AUs and FEs jointly. A shared CLIP vision encoder with MoE layers first extracts frame-wise features from FE or AU videos. Then, these features are fed into two Transformer-based temporal models, $\Phi_{\mathrm{exp}}(\cdot)$ and $\Phi_{\mathrm{au}}(\cdot)$, for task-specific temporal modeling. The DFER branch aggregates sequential features into a clip-level representation $\boldsymbol{Z}^{\mathrm{exp}}$. In contrast, the AU branch performs temporal interaction among the input frames and selects the temporally enhanced feature of the center frame as the AU representation $\boldsymbol{Z}^{\mathrm{au}}$. Finally,  $\boldsymbol{Z}^{\mathrm{exp}}$ and $\boldsymbol{Z}^{\mathrm{au}}$ are  fed into independent linear heads for classification. This Baseline model corresponds to the visual backbone in Fig.~\ref{fig:oview}.}
    \label{baseline_image}
\end{figure}

Our Baseline, a multi-task model, consists of a shared visual backbone and two independent linear layers. As illustrated in Fig.~\ref{baseline_image}, the shared visual backbone extracts unified dynamic facial representations from heterogeneous video data. It includes a shared CLIP vision encoder with MoE (Mixture of Experts) layers\footnote{\emph{Directly sharing the original CLIP vision encoder makes it difficult to learn both tasks effectively according to our experiments. Therefore, we insert MoE layers to enable joint learning of the two tasks,  following the mainstream practice \cite{chen2025static,dai2024deepseekmoe,jain2023damex,chen2023adamv}. Details are provided in Sec. E of the supplementary material.}}, denoted as $E_v(\cdot)$, and two task-specific temporal modules: an expression temporal model $\Phi_{\mathrm{exp}}(\cdot)$ and an AU temporal model $\Phi_{\mathrm{au}}(\cdot)$. Both temporal modules are built with standard Transformer blocks.

Specifically, given an expression video sequence $\{\boldsymbol{v}_1^{\mathrm{exp}}, \boldsymbol{v}_2^{\mathrm{exp}}, \ldots, \boldsymbol{v}_n^{\mathrm{exp}}\}$ and an AU video sequence $\{\boldsymbol{v}_1^{\mathrm{au}}, \boldsymbol{v}_2^{\mathrm{au}}, \ldots, \boldsymbol{v}_n^{\mathrm{au}}\}$, the shared vision encoder first extracts visual features as:
\begin{equation}
\{\boldsymbol{f}_1^{\mathrm{exp}}, \boldsymbol{f}_2^{\mathrm{exp}}, \ldots, \boldsymbol{f}_n^{\mathrm{exp}}\}
=
E_v(\{\boldsymbol{v}_1^{\mathrm{exp}}, \boldsymbol{v}_2^{\mathrm{exp}}, \ldots, \boldsymbol{v}_n^{\mathrm{exp}}\}),
\end{equation}
\begin{equation}
\{\boldsymbol{f}_1^{\mathrm{au}}, \boldsymbol{f}_2^{\mathrm{au}}, \ldots, \boldsymbol{f}_n^{\mathrm{au}}\}
=
E_v(\{\boldsymbol{v}_1^{\mathrm{au}}, \boldsymbol{v}_2^{\mathrm{au}}, \ldots, \boldsymbol{v}_n^{\mathrm{au}}\}).
\end{equation}

The frame-level features are then fed into their corresponding temporal modules, yielding the task-specific representations for DFER and AU detection, respectively:
\begin{equation}
\boldsymbol{Z}^{\mathrm{exp}} = \Phi_{\mathrm{exp}}(\{\boldsymbol{f}_1^{\mathrm{exp}}, \boldsymbol{f}_2^{\mathrm{exp}}, \ldots, \boldsymbol{f}_n^{\mathrm{exp}}\}), 
\end{equation}
\begin{equation}
\{\tilde{\boldsymbol{Z}}_1^{\mathrm{au}}, \tilde{\boldsymbol{Z}}_2^{\mathrm{au}}, \ldots, \tilde{\boldsymbol{Z}}_n^{\mathrm{au}}\}
=
\Phi_{\mathrm{au}}(\{\boldsymbol{f}_1^{\mathrm{au}}, \boldsymbol{f}_2^{\mathrm{au}}, \ldots, \boldsymbol{f}_n^{\mathrm{au}}\}),
\end{equation}
\begin{equation}
\boldsymbol{Z}^{\mathrm{au}}=\tilde{\boldsymbol{Z}}_t^{\mathrm{au}}, \quad t = \lfloor \frac{n}{2} \rfloor,
\end{equation}
where $t$ denotes the index of the center frame in the video clip. We use the temporally enhanced feature of the center frame to predict its AU activations, while the remaining frames provide temporal context.

Finally, the task-specific representations are mapped to prediction logits through their corresponding classification heads:
\begin{equation}
\boldsymbol{o}^{\mathrm{exp}} = \boldsymbol{W}_{\mathrm{cls}}^{\mathrm{exp}} \boldsymbol{Z}^{\mathrm{exp}} + \boldsymbol{b}_{\mathrm{cls}}^{\mathrm{exp}}, \qquad
\boldsymbol{o}^{\mathrm{au}} = \boldsymbol{W}_{\mathrm{cls}}^{\mathrm{au}} \boldsymbol{Z}^{\mathrm{au}} + \boldsymbol{b}_{\mathrm{cls}}^{\mathrm{au}},
\end{equation}
where $\boldsymbol{o}^{\mathrm{exp}} \in \mathbb{R}^{K}$ denotes the prediction over the $K$ expression categories for the DFER task, and $\boldsymbol{o}^{\mathrm{au}} \in \mathbb{R}^{M}$ denotes the prediction over the $M$ AU labels for the AU detection task. $\boldsymbol{W}_{\mathrm{cls}}^{\mathrm{exp}}$ and $\boldsymbol{W}_{\mathrm{cls}}^{\mathrm{au}}$ denote the weight matrices of the linear classification heads for DFER and AU detection. $\boldsymbol{b}_{\mathrm{cls}}^{\mathrm{exp}}$ and $\boldsymbol{b}_{\mathrm{cls}}^{\mathrm{au}}$ denote the corresponding bias terms.

DFER is a single-label multi-class classification task, thus the softmax cross-entropy loss is adopted:
\begin{equation}
\mathcal{L}_{\mathrm{dfe}}
=
-\frac{1}{B}
\sum_{m=1}^{B}
\sum_{n=1}^{K}
y_{m,n}
\log
\frac{\exp\left(\boldsymbol{o}^{\mathrm{exp}}_{m,n}\right)}
{\sum_{j=1}^{K}\exp\left(\boldsymbol{o}^{\mathrm{exp}}_{m,j}\right)},
\end{equation}
where $B$ denotes the batch size, and $y_{m,n} \in \{0,1\}$ indicates the ground-truth label of the $m$-th sample for the $n$-th expression category, satisfying $\sum_{n=1}^{K} y_{m,n} = 1$.

AU detection is a multi-label binary classification task, thus the binary cross-entropy loss is employed:
\begin{equation}
\begin{aligned}
\mathcal{L}_{\mathrm{au}} ={}& -\frac{1}{B}\sum_{m=1}^{B}\sum_{n=1}^{M}
\Big[
y_{m,n}\log \sigma\left(\boldsymbol{o}^{\mathrm{au}}_{m,n}\right) \\
&\qquad\qquad\quad
+ \left(1-y_{m,n}\right)\log\left(1-\sigma\left(\boldsymbol{o}^{\mathrm{au}}_{m,n}\right)\right)
\Big],
\end{aligned}
\end{equation}
where $y_{m,n} \in \{0,1\}$ indicates whether the $n$-th AU is activated in the $m$-th sample, and $\sigma(\cdot)$ denotes the sigmoid function.

\subsubsection{CLIP-Style Prompt Learning}

\noindent Vision language models, represented by CLIP \cite{radford2021learning}, achieve cross-modal alignment through large-scale image–text contrastive learning. Given a set of images and class labels, i.e. \(\boldsymbol{I}\) and \(y\), by first constructing a textual description \(\boldsymbol{T}_y\) for the label \(y\), CLIP formulates the classification task as matching the similarity between the image feature \(\boldsymbol{f}_I = E_I(\boldsymbol{I})\) and the text feature \(\boldsymbol{f}_T = E_T(\boldsymbol{T}_y)\):

\begin{equation}
p\left(y\middle| \boldsymbol{I}\right)
= \frac{\exp\!\left(\cos\!\left(\boldsymbol{f}_I, \boldsymbol{f}_{T_y}\right) / \tau\right)}
       {\sum_{i=1}^{y} \exp\!\left(\cos\!\left(\boldsymbol{f}_I, \boldsymbol{f}_{T_i}\right) / \tau\right)},
\end{equation}
where \(E_I\) and \(E_T\) denote the image and text encoders respectively, and \(\tau\) is the temperature hyperparameter. 

Building on this formulation, CoOp \cite{zhou2022learning} further introduces learnable context vectors \(\boldsymbol{v} = (v_1, v_2, \dots, v_M)\), expanding the textual representation of a class to
\begin{equation}
\boldsymbol{T}_y = \left[v_1, v_2, \ldots, v_M,\ \text{\texttt{"class name of }y\texttt{"}}\right],
\end{equation}
which enables the model to automatically adapt to the task context and to optimize the prompt representation. Therefore, we continue to follow this scheme in our method.

\subsection{Framework Overview}

As illustrated in Fig.~\ref{fig:oview}, the proposed SSM framework is built upon the Baseline introduced in Sec.~\ref{sec:Baseline}. SSM retains the same shared visual backbone. It produces the task-specific visual representations \(\boldsymbol{Z}^{\text{exp}}\) for DFER and \(\boldsymbol{Z}^{\text{au}}\) for AU detection.

Different from the Baseline model, which performs classification using two independent linear heads, SSM reformulates both DFER and AU detection within a unified vision-text alignment  space. In this space, predictions are made by measuring the similarity between task-specific visual features and the corresponding textual embeddings.

In the textual domain, we do not rely on bare class names. Instead, we construct \(K\) expression-related natural language descriptions and \(M\) facial-action natural language descriptions based on FACS knowledge. Here, \(K\) and \(M\) denote the numbers of classes for DFER and AU detection, respectively. The AU semantic descriptions serve as the basic units for composing the dynamic expression text prompts, i.e., \(\boldsymbol{t}^{\text{au}} \subset \boldsymbol{t}^{\text{exp}}\). Here, \(\boldsymbol{t}^{\text{exp}} \in \mathbb{R}^{1 \times K}\) and \(\boldsymbol{t}^{\text{au}} \in \mathbb{R}^{1 \times M}\). After encoding with the shared CLIP text encoder \(E_t(\cdot)\), the two sets of textual descriptions become
\begin{equation}
\boldsymbol{T}^{\text{exp}} = E_t(\boldsymbol{t}^{\text{exp}}) \in \mathbb{R}^{K \times d}, \qquad
\boldsymbol{T}^{\text{au}} = E_t(\boldsymbol{t}^{\text{au}}) \in \mathbb{R}^{M \times d},
\end{equation}
where \(d\) is the dimensionality of the encoded text embeddings. 
Finally, we perform joint text-driven classification training for both tasks. Concretely, the DFER loss is defined as
\begin{equation}
\mathcal{L}_\text{dfe}
= -\frac{1}{B}\sum_{m=1}^{B}\sum_{n=1}^{K} 
{y_{m,n}}\log\frac{\bigl(\exp\bigl(\,\boldsymbol{Z}_m^\text{exp}\cdot \boldsymbol{T}_n^\text{exp}/\tau\,\bigr)\bigr)}
{\displaystyle\sum_{j=1}^{K}\exp\bigl(\boldsymbol{Z}_m^\text{exp}\cdot \boldsymbol{T}_j^\text{exp}/\tau\bigr)},
\end{equation}
where $y_{m,n} \in \{0,1\}$ denotes the ground-truth label of the $m$-th sample for the $n$-th expression category, and $\sum_{n=1}^{K} y_{m,n} = 1$.

The AU detection loss is given by the average binary cross-entropy over the \(M\) AUs:
\begin{equation}\label{eq:au_loss}
\begin{aligned}
\mathcal{L}_\text{au} ={}& -\frac{1}{B}\sum_{m=1}^{B}\sum_{n=1}^{M}
\Big(y_{m,n} \log\big(\sigma(\boldsymbol{Z}_m^\text{au}\!\cdot\! \boldsymbol{T}_n^\text{au}/\tau)\big) \\
&\qquad\qquad\quad
+ (1-y_{m,n})\log\big(1-\sigma(\boldsymbol{Z}_m^\text{au}\!\cdot\! \boldsymbol{T}_n^\text{au}/\tau)\big) \Big),
\end{aligned}
\end{equation}
where $y_{m,n} \in \{0,1\}$ denotes whether the $n$-th AU is activated in the $m$-th sample.

The total loss for joint training is then
\begin{equation}
\mathcal{L}_{\mathrm{total}} \;=\; \frac{1}{1+\lambda\,}\mathcal{L}_\text{dfe} \;+\; \frac{\lambda\,}{1+\lambda\,}\mathcal{L}_\text{au},
\label{eq:lamda}
\end{equation}
with \(\lambda\) as a task-balancing hyperparameter.

\subsection{Textual Semantic Prototype Module}

This subsection introduces how task-specific textual descriptions are constructed in the Textual Semantic Prototype (TSP) module.

As shown in the left part of Fig.~\ref{fig:oview}, we first construct fixed text templates for the tasks based on FACS knowledge~\cite{ekman2002facs}.
For AU detection, each template uses the canonical AU description in the FACS Manual~\cite{ekman2002facs}. For FEs, we compose each template according to the prototypical AU–expression configurations in the FACS Investigator’s Guide~\cite{ekman2002facs}, retaining only AUs available in the paired AU dataset. These combinations are dataset-constrained approximations rather than original FACS prototypes. For non-basic expression categories annotated in some dataset like MAFW \cite{liu2022mafw} not covered by the Guide, we use manually composed, dataset-constrained AU descriptions as initial semantic approximations\footnote{\emph{The complete templates are listed in Table S4 in Sec. D of the supplementary material.}}.

These templates are denoted as $d_{\mathrm{exp}}^{k}$ and $d_{\mathrm{au}}^{m}$, where $k \in \{1,2,\ldots,K\}$ and $m \in \{1,2,\ldots,M\}$. We then map them into token sequences through a tokenizer:
\begin{equation}
d_{\mathrm{exp}}^{k}=\{w_{1}^\mathrm{exp},w_{2}^\mathrm{exp},\ldots,w_{l}^\mathrm{exp}\},
\end{equation}
\begin{equation}
d_{\mathrm{au}}^{m}=\{w_{1}^\mathrm{au},w_{2}^\mathrm{au},\ldots,w_{l}^\mathrm{au}\}.
\end{equation}

Thus, the final text descriptions can be expressed as:
\begin{equation}
\boldsymbol{t}^{\mathrm{exp}}=[ p_{1}^\mathrm{exp},p_{2}^\mathrm{exp},\ldots,p_{c}^\mathrm{exp},\,d_{\mathrm{exp}}],
\end{equation}
\begin{equation}
\boldsymbol{t}^{\mathrm{au}}= [p_{1}^\mathrm{au},p_{2}^\mathrm{au},\ldots,p_{c}^\mathrm{au},\,d_{\mathrm{au}}],
\end{equation}
where $[p_{1}^\mathrm{exp},\ldots,p_{c}^\mathrm{exp}]$ and $[p_{1}^\mathrm{au},\ldots,p_{c}^\mathrm{au}]$ denote learnable context prompts.

However, in CLIP-style classification, textual features are usually constructed as independent class prototypes. Such independently constructed prototypes cannot explicitly model the semantic relationships between AUs and facial expression categories. Therefore, as shown in Fig.~\ref{dpm_image}, we further design the Dynamic Prior Mapping (DPM) module to learn the semantic dependencies between the two tasks through bidirectional prior mapping matrices. DPM serves as a bridge in the high-level textual semantic space and enables cross-task interaction between AU and FE prototypes.

\subsection{Dynamic Prior Mapping Module}

\begin{figure}[t]  
    \centering
    \includegraphics[width=\columnwidth]{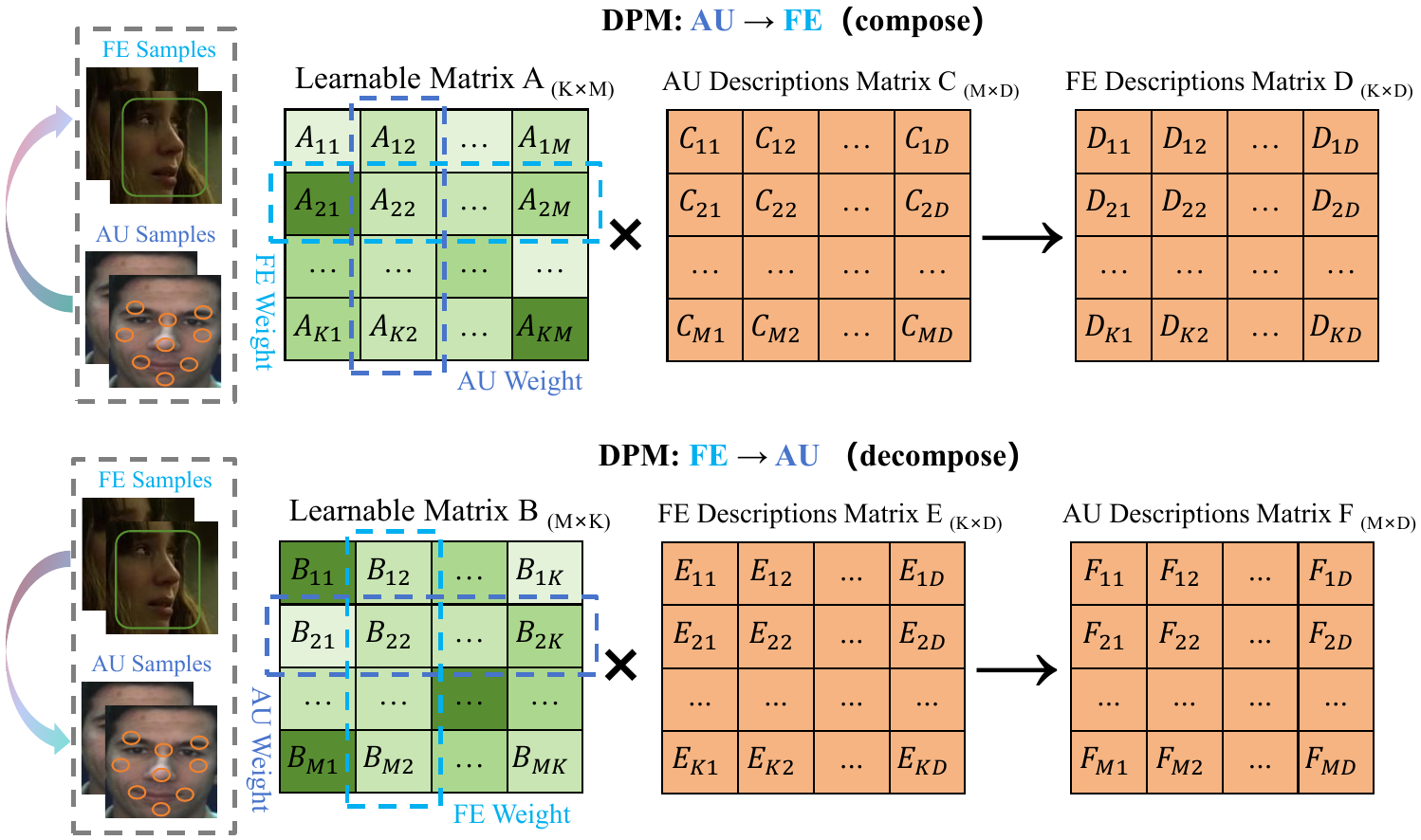}
    \caption{The linear feature transformation in the DPM module is to multiply a learnable weight matrix with a task-specific semantic matrix. Composition transformation combines the semantics of multiple AUs into the semantics of a single FE. Hence, the AU semantic matrix is transformed into a new FE semantic matrix. Decomposition transformation  decomposes the semantics of a single FE into the semantics of multiple AUs. Hence, the FE semantic matrix is transformed into a new AU semantic matrix. Matrices $\boldsymbol{A}$, $\boldsymbol{B}$, $\boldsymbol{C}$, $\boldsymbol{D}$, $\boldsymbol{E}$, and $\boldsymbol{F}$ denote
$\boldsymbol{W}^{\text{au}\rightarrow\text{exp}}$,
$\boldsymbol{W}^{\text{exp}\rightarrow\text{au}}$,
$\boldsymbol{T}^{\text{au}}$,
$\tilde{\boldsymbol{T}}^{\text{exp}}$,
$\boldsymbol{T}^{\text{exp}}$, and
$\tilde{\boldsymbol{T}}^{\text{au}}$, respectively.}
    \label{dpm_image}
\end{figure}

We propose Dynamic Prior Mapping (DPM), a learnable, bidirectional, and differentiable mapping mechanism in the textual semantic space. Its main objective is to establish a dynamic correspondence bridge between local AU semantics and global FE semantics. This design enhances discriminability and implicitly mitigates dataset bias. 

Unlike fixed relation matrices, DPM operates on textual prototypes and updates the two mapping directions independently. FACS provides only the initialization.

As illustrated in Fig.~\ref{dpm_image}, the DPM module consists of two learnable mapping matrices: \(\boldsymbol{W}^{\text{au}\rightarrow\text{exp}} \in \mathbb{R}^{K \times M}\) and \(\boldsymbol{W}^{\text{exp}\rightarrow\text{au}} \in \mathbb{R}^{M \times K}\). These matrices are initialized from FACS-derived priors and  model the semantic mappings \(\text{AU detection}\!\to\!\text{DFER}\) and \(\text{DFER}\!\to\!\text{AU detection}\).

Specifically, we construct a binary prior matrix \(\boldsymbol{P} \in \mathbb{R}^{K \times M}\) from the same
FACS-derived AU--expression configurations used in TSP. Only AUs annotated in the paired AU dataset are retained. Here, \(P_{k,m} = 1\) if AU \(m\) belongs to the selected AU set for expression category \(k\), and \(P_{k,m} = 0\) otherwise.


To avoid overly hard constraints, we normalize the prior matrix
row-wise and use it to initialize the learnable mapping matrix as:
\begin{equation}
\boldsymbol{W}^{\text{au}\rightarrow\text{exp}}_{0}
=
\mathrm{Normalize}_{\mathrm{row}}(\boldsymbol{P}),
\end{equation}
the reverse-direction matrix is initialized as its transpose:
\begin{equation}
\boldsymbol{W}^{\text{exp}\rightarrow\text{au}}_{0}
=
\mathrm{Normalize}_{\mathrm{row}}(\boldsymbol{P}^{\top}).
\end{equation}

During training, the two matrices are updated independently through backpropagation. This allows the two directions to learn asymmetric relations and adapt to the statistics of each dataset pair. Thus, the FACS prior provides the initialization, while the final mappings are learned from the data.

Given the DFER textual embedding matrix \(\boldsymbol{T}^\text{exp}\in\mathbb{R}^{K\times d}\) and the AU textual embedding matrix \(\boldsymbol{T}^\text{au}\in\mathbb{R}^{M\times d}\), the DPM bidirectional mappings are defined as:
\begin{equation}
\tilde{\boldsymbol{T}}^{\text{exp}} =
\text{Softmax}_{\text{row}}
\left(
\boldsymbol{W}^{\text{au}\rightarrow\text{exp}
}/\tau_m
\right)
\boldsymbol{T}^{\text{au}}, 
\end{equation}
\begin{equation}
\tilde{\boldsymbol{T}}^{\text{au}} =
\text{Softmax}_{\text{row}}
\left(
\boldsymbol{W}^{\text{exp}\rightarrow\text{au}
}/\tau_m
\right)
\boldsymbol{T}^{\text{exp}},
\end{equation}
where \(\operatorname{Softmax}_{\text{row}}(\cdot)\) refers to a row-wise softmax, temperature-scaled by \(\tau_m\), to ensure numerical stability and introduce a non-linear normalization over the mapping weights. This normalization makes the associations more interpretable. Here, \(\tilde{\boldsymbol{T}}^{\text{exp}}\) is the expression-semantic mapping generated from AU descriptions, whereas \(\tilde{\boldsymbol{T}}^{\text{au}}\) is the reverse mapping from expression descriptions to AU semantics. Through this bidirectional association, our model explicitly captures the complementary relationships between the AU and DFER tasks.

The final semantically enhanced representations for DFER and AU detection tasks are obtained through residual-style updates:
\begin{equation}
\boldsymbol{T}^{\text{exp}}
=
\boldsymbol{T}^{\text{exp}}
+ \alpha \tilde{\boldsymbol{T}}^{\text{exp}}, 
\label{eq:alpha}
\end{equation}
\begin{equation}
\boldsymbol{T}^{\text{au}}
=
\boldsymbol{T}^{\text{au}}
+ \beta \tilde{\boldsymbol{T}}^{\text{au}},
\label{eq:beta}
\end{equation}
where \(\alpha\) and \(\beta\) are two learnable weighting factors.

 The cosine similarities between the visual features and all candidate textual prototypes are first computed as
\begin{equation}
S(\boldsymbol{Z}, \boldsymbol{T}) = \frac{\boldsymbol{Z} \cdot \boldsymbol{T}}{\|\boldsymbol{Z}\| \, \|\boldsymbol{T}\|},
\end{equation}
where $\boldsymbol{Z}$ and $\boldsymbol{T}$ denote the visual and textual feature vectors, respectively. The final predictions are then obtained by selecting the prototype with the highest similarity score for DFER, while for AU detection, the similarity scores are used as confidence values for each AU.


\section{Experiments}
\subsection{Datasets}
We evaluate the proposed method on two laboratory dynamic AU detection datasets, BP4D \cite{zhang2014bp4d} and DISFA \cite{mavadati2013disfa}, and three in-the-wild dynamic facial expression recognition datasets, DFEW \cite{jiang2020dfew}, MAFW \cite{liu2022mafw}, and FERV39K \cite{wang2022ferv39k}. All of them are publicly available and widely used benchmarks. For each dataset, we follow the official split protocol. For  AU detection, we use F1 score as the evaluation metric. For DFER, following prior studies \cite{ma2023logo,li2023intensity,liu2023expression,zhao2021former}, we use Unweighted Average Recall (UAR) and Weighted Average Recall (WAR) as the evaluation metrics.

\subsubsection{AU Datasets}
\textbf{BP4D} \cite{zhang2014bp4d} is a laboratory-collected 3D dynamic spontaneous facial expression dataset with 41 subjects and 328 high-resolution videos. Twelve AUs are annotated at the frame level, yielding approximately 146,000 labeled frames. The dataset follows a 3-fold cross-validation protocol. 
\textbf{DISFA} \cite{mavadati2013disfa} is a laboratory-collected dynamic facial expression dataset with 27 subjects and approximately 130,000 frames. Twelve AUs are annotated at the frame level with intensity levels from 0 to 5. Following common practice \cite{yang2021exploiting}, we select eight AUs for activation detection and adopt a 3-fold cross-validation protocol.

\subsubsection{DFER Datasets}
\textbf{DFEW} \cite{jiang2020dfew} is a large-scale in-the-wild dynamic facial expression dataset with 16,372 video clips collected from approximately 1,500 movies. It covers seven basic expression categories and follows a 5-fold cross-validation protocol. Each clip is annotated at the video level by multiple annotators to ensure label reliability. 
\textbf{MAFW} \cite{liu2022mafw} is a large-scale in-the-wild multimodal (video-audio) compound emotion dataset with 10,045 clips and 11 emotion categories. Each clip is annotated at the video level by multiple annotators. The dataset provides both single-label and multi-label splits. The dataset follows a 5-fold cross-validation protocol. 
\textbf{FERV39K} \cite{wang2022ferv39k} is a large-scale multi-scene in-the-wild video expression recognition dataset with 38,935 clips. It covers seven basic expression categories across diverse scene types. It adopts the official train/test split, with 31,088 clips for training and 7,847 for testing.


\subsection{Implementation Details}

Facial images are aligned and cropped to a resolution of 224$\times$224. Data augmentation includes random cropping, random erasing, horizontal flipping, and color jittering. The encoder is based on CLIP-ViT-B/16 \cite{radford2021learning}. MoE layers are inserted into the FFNs of the last six layers of the CLIP vision encoder, whose  top-$k$ is set to 2 and the number of private experts is set to 4. The input and output dimensions of MoE layers are both 768, and the hidden feature dimension is 512. For the DFER task, following previous works \cite{li2023intensity,zhao2021former,zhao2023prompting,chen2024static}, we uniformly sample video clips. Each sample contains 16 frames. For the temporal model $\Phi_{\mathrm{exp}}(\cdot)$, the numbers of Transformer layers and attention heads are set to 1 and 8 by default, respectively, to avoid overfitting. For the AU detection task, the video sampling strategy and the hyperparameters of the temporal model $\Phi_{\mathrm{au}}(\cdot)$ are kept consistent with those of the DFER task. On the text side, we adopt the CoOp design \cite{zhou2022learning}. It includes 8 learnable context tokens and a fixed textual template. By default, the fixed textual template is placed after the learnable context tokens. In addition, the loss weighting coefficient $\lambda$ in Eqn. \ref{eq:lamda} is set to 2 by default to balance the losses between tasks. The initial values of $\alpha$ and $\beta$ in Eqn. \ref{eq:alpha} and Eqn. \ref{eq:beta} are both set to 0.1. The temperature hyperparameters $\tau$ and $\tau_m$ are both set to 0.01.

During training, the AdamW optimizer is used. The learning rate for the visual encoder branch is set to $1\times10^{-6}$. The learning rate for the remaining branches is set to $1\times10^{-4}$. The weight decay is uniformly set to $1\times10^{-4}$. We adopt a multi-step decay schedule. The learning rate of all components is reduced to 0.1 times the previous value every 10 epochs. The DFER and AU detection mini-batches contain 12 and 8 clips, respectively, and each clip contains 16 frames. At each optimization step, the model separately processes one DFER mini-batch and one AU detection mini-batch, and the two task losses are combined for a joint update. The longer data loader determines the number of steps in each epoch, while the shorter loader is restarted after exhaustion. Each DFER clip has one clip-level expression label. The AU detection datasets provide frame-level annotations for the video frames. For each AU detection sample, the loss is computed using the AU label of the center frame, while the remaining frames provide temporal visual context. Joint training is performed for 30 epochs. The random seed is set to 1 for all experiments to ensure reproducibility. All experiments are conducted on 8 NVIDIA 4090 GPUs.\footnote{\emph{The sensitivity to the task-loss weight and the computational complexity are analyzed in Secs. F and I of the supplementary material, respectively.}}

Additionally, to verify the advantage of our framework, we also trained a single-task learning model, which is referred to as STL. The visual backbone is also the standard CLIP-ViT-B/16 \cite{radford2021learning}, and the two tasks are trained separately, with a temporal module and a linear layer for each task. Other model configurations are kept the same.

\begin{table*}[!t]
  \centering
  \caption{F1 scores over 12 AUs on the BP4D dataset, using joint learning based on the DFEW dataset. The best results are highlighted in bold, and the second-best underlined. STL refers to the single-task counterpart.}
  \label{tab:bp4d_results}
  \begin{adjustbox}{max width=\textwidth}
  \small
  \begin{tabular}{l l c c c c c c c c c c c c|c}
    \toprule
    Methods & Backbone &
      AU1 & AU2 & AU4 & AU6 & AU7 & 
      AU10 & AU12 & AU14 & AU15 & AU17 & AU23 & AU24 &
      Avg \\
    \midrule
    KSRL\cite{chang2022knowledge}        & ResNet-50          & 53.3 & 47.4 & 56.2 & 79.4 & 80.7 & 85.1 & 89.0 & 67.4 & 55.9 & 61.9 & 48.5 & 49.0 & 64.5 \\
    KS\cite{li2023knowledge}          & ResNet-18             & 55.3 & 48.6 & 57.1 & 77.5 & \underline{81.8} & 83.3 & 86.4 & 62.8 & 52.3 & 61.3 & 51.6 & 58.3 & 64.7 \\
    MDHR\cite{wang2024multi}        & Swin-B                  & 58.3 & 50.9 & 58.9 & 78.4 & 80.3 & 84.9 & 88.2 & 69.5 & 56.0 & 65.5 & 49.5 & 59.3 & 66.6 \\
    CLEF\cite{zhang2023weakly}        & CLIP-ViT-B/16         & 55.8 & 46.8 & 63.3 & 79.5 & 77.6 & 83.6 & 87.8 & 67.3 & 55.2 & 63.5 & 53.0 & 57.8 & 65.9 \\
    AUFormer\cite{yuan2024auformer}    & ViT-B/16             & - & - & - & - & - & - & - & - & - & - & - & - & 66.2 \\
    FMAE\cite{ning2024representation}        & ViT-L/16       & 59.2 & 50.0 & 62.7 & 80.0 & 79.2 & 84.7 & \underline{89.8} & 63.5 & 52.8 & 65.1 & \underline{55.3} & 56.9 & 66.6\\
    FMAE-IAT\cite{ning2024representation}    & ViT-L/16       & \underline{62.7} & \underline{51.9} & 62.7 & 79.8 & 80.1 & 84.8 & \textbf{89.9} & 64.6 & 54.9 & 65.4 & 53.1 & 54.7 & 67.1 \\
    MAE-Face\cite{ma2024facial}    & ViT-B/16                 & 62.5 & \textbf{56.4} & \textbf{66.3} & 79.6 & 79.6 & \underline{85.6} & 89.1 & 64.2 & 54.5 & 65.0 & 53.8 & 51.8 & 67.4 \\
    AU-TTT\cite{xing2025ttt} & ViT-S/16 & - & - & - & - & - & - & - & - & - & - & - & - & 65.6 \\
    FLCM\cite{huang2025facial} & ResNet-50 & 60.6 & 50.3 & 64.2 & \underline{80.7} & 80.5 & \textbf{85.9} & 88.6 & 68.0 & \underline{57.3} & 63.4 & 52.0 & \textbf{60.5} & \underline{67.7} \\
    HiVA\cite{li2026hierarchical} & Swin-B & 54.3 & 49.7 & 63.3 & 79.3 & 79.8 & 84.5 & 88.8 & 68.5 & 57.0 & 62.6 & 53.1 & 56.8 & 66.5 \\
    CausalAffect (+EmotioNet)\cite{hu2025causalaffect} & ResNet-50 & \textbf{67.1} & 43.6 & \underline{66.0} & 80.1 & 79.1 & 84.8 & 88.9 & \textbf{71.1} & 55.6 & \underline{66.6} & 47.5 & 58.8 & 67.4 \\
    \midrule

      \textbf{STL (Ours)}   & CLIP-ViT-B/16      & 58.8 & 48.0 & 60.5 & 78.4 & 79.6 & 84.3 & 88.8 & 69.5 & 51.7 & 65.7 & 53.1 & 55.7 & 66.2 \\    
     
      \textbf{SSM (Ours)}  & CLIP-ViT-B/16      & 61.0 & 48.4 & 56.0 & \textbf{81.8} & \textbf{83.1} & 84.9 & 88.7 & \underline{70.7} & \textbf{59.5} & \textbf{68.0} & \textbf{60.1} & \underline{59.8} & \textbf{68.5} \\
    \bottomrule
  \end{tabular}
  \end{adjustbox}
\end{table*}

\begin{table*}[!t]
  \centering
  \caption{F1 scores over 8 AUs on the DISFA dataset, using joint learning based on the DFEW dataset. The best results are highlighted in bold, and the second-best underlined. STL refers to the single-task counterpart.}
  \label{tab:disfa_results}
  \begin{adjustbox}{max width=\textwidth}
  \small
  \begin{tabular}{l l c c c c c c c c|c}
    \toprule
    Methods & Backbone &
      AU1 & AU2 & AU4 & AU6 &
      AU9 & AU12 & AU25 & AU26 &
      Avg \\
    \midrule
    KSRL\cite{chang2022knowledge}        & ResNet-50            & 60.4 & 59.2 & 67.5 & 52.7 & 51.5 & 76.1 & 91.3 & 57.7 & 64.5 \\
    KS\cite{li2023knowledge}          & ResNet-18            & 53.8 & 59.9 & 69.2 & 54.2 & 50.8 & 75.8 & 92.2 & 46.8 & 62.8 \\
    MDHR\cite{wang2024multi}        & Swin-B               & 65.4 & 60.2 & 75.2 & 50.2 & 52.4 & 74.3 & 93.7 & 58.2 & 66.2 \\
    CLEF\cite{zhang2023weakly}        & CLIP-ViT-B/16        & 64.3 & 61.8 & 68.4 & 49.0 & 55.2 & 72.9 & 89.9 & 57.0 & 64.8 \\
    AUFormer\cite{yuan2024auformer}    & ViT-B/16             & --   & --   & --   & --   & --   & --   & --   & --   & 66.4 \\
    FMAE\cite{ning2024representation}        & ViT-L/16             & 62.7 & 59.5 & 67.3 & 55.6 & \underline{61.8} & 77.9 & 95.0 & 69.8 & 68.7 \\
    FMAE-IAT\cite{ning2024representation}    & ViT-L/16             & 64.7 & 61.3 & 70.8 & 58.1 & 59.4 & \textbf{79.9} & 95.2 & \underline{71.3} & 70.1 \\
    MAE-Face\cite{ma2024facial}    & ViT-B/16             & \underline{68.4} & 59.4 & \underline{76.5} & \underline{58.4} & 56.7 & 78.5 & \textbf{96.6} & \textbf{71.7} & 70.8 \\
    AU-TTT\cite{xing2025ttt} & ViT-S/16 & - & - & - & - & - & - & - & - & 66.4 \\
    FLCM\cite{huang2025facial} & ResNet-50 & 59.3 & 62.1 & 73.7 & 55.3 & 56.3 & 79.1 & 93.9 & 62.4 & 67.8 \\
    HiVA\cite{li2026hierarchical} & Swin-B & 60.6 & 58.4 & 75.4 & 51.0 & 61.2 & 74.8 & 93.9 & 63.8 & 67.4 \\
    CausalAffect (+EmotioNet)\cite{hu2025causalaffect} & ResNet-50 & 68.1 & 63.2 & \textbf{77.6} & \textbf{64.1} & \textbf{74.0} & 69.3 & 83.7 & 68.7 & \underline{71.1} \\

    \midrule
      \textbf{STL (Ours)}   & CLIP-ViT-B/16     & 61.4 & \underline{70.9} & 69.8 & 57.1 & 56.0 & 77.3 & \underline{95.8} & 68.7 & 69.6 \\
      \textbf{SSM (Ours)}  & CLIP-ViT-B/16     & \textbf{68.6} & \textbf{74.6} & 73.9 & 56.1 & 57.6 & \underline{79.4} & 95.6 & 69.8 & \textbf{71.9} \\
    \bottomrule
  \end{tabular}
  \end{adjustbox}
\end{table*}

\begin{table*}[!t]
    \centering
    \caption{Comparison with state-of-the-art methods on DFEW, FERV39K, and MAFW. UAR: Unweighted Average Recall; WAR: Weighted Average Recall. Using joint learning based on the BP4D dataset. The best results are highlighted in bold, and the second-best underlined. STL refers to the single-task counterpart.}
    \label{tab:dfer}
    \small
    \begin{tabular}{lccccccc}
        \toprule
        \multirow{2}{*}{Method}                    & \multirow{2}{*}{Backbone} & \multicolumn{2}{c}{DFEW} & \multicolumn{2}{c}{FERV39K} & \multicolumn{2}{c}{MAFW}                                                             \\
        \cline{3-4} \cline{5-6} \cline{7-8}
                                                   &                           & UAR                      & WAR                         & UAR                      & WAR               & UAR               & WAR               \\
        \hline
        \textbf{\textit{Supervised learning models}} & & & & & & &\\
        Former-DFER\cite{zhao2021former}         & Transformer               & 53.69                    & 65.70                       & 37.20                    & 46.85             & 31.16             & 43.27             \\
        NR-DFERNet\cite{li2022nr}                & CNN-Transformer           & 54.21                    & 68.19                       & 33.99                    & 45.97             & -                 & -                 \\
    
        EST\cite{liu2023expression}               & ResNet-18                 & 53.43                    & 65.85                       & -                        & -                 & -                 & -                 \\
        Freq-HD\cite{tao2023freq}                & VGG13-LSTM                & 46.85                    & 55.68                       & 33.07                    & 45.26             & -                 & -                 \\
        IAL\cite{li2023intensity}                 & ResNet-18                 & 55.71                    & 69.24                       & 35.82                    & 48.54             & -                 & -                 \\
       
        M3DFEL\cite{wang2023rethinking}           & ResNet-18-3D              & 56.10                    & 69.25                       & 35.94                    & 47.67             & -                 & -                 \\
        IFDD-3DViT\cite{wang2025lifting}        & ViT-B/16                  & 61.19                    & 73.82                      & 39.15                    &  51.09             & 39.31 & 53.92             \\
        \hline
        \textbf{\textit{Self-supervised learning models}} & & & & & & &\\
        SVFAP\cite{sun2024svfap}                 & ViT-B/16                  & 62.83                    & 74.27                       & 42.14                    & 52.29             & 41.19             & 54.28   
        \\
        MAE-DFER\cite{sun2023mae}               & ViT-B/16                  & 63.41                    & 74.43                       & 43.12                    & 52.07             & 41.62             & 54.31             \\
        \hline
        \textbf{\textit{Vision-language models}} & & & & & & & \\
        CLIPER\cite{li2024cliper}                 & CLIP-ViT-B/16             & 57.56                    & 70.84                       & 41.23                    & 51.34             & -                 & -                 \\
        EmoCLIP\cite{foteinopoulou2024emoclip}  & CLIP-ViT-B/32             & 58.04                    & 62.12                       & 31.41                    & 36.18             & 34.24             & 41.46             \\
        DFER-CLIP\cite{zhao2023prompting}         & CLIP-ViT-B/32             & 59.61                    & 71.25                       & 41.27                    & 51.65             & 39.89             & 52.55             \\

        DFLM\cite{han2024dflm}          & CLIP-ViT-B/32             & 59.77        & 71.40                       & 41.25                    & 51.31             & 41.23             & 53.65             \\
        CLIP-Guided-DFER\cite{zhang2024clip}          & CLIP-ViT-B/32             & 60.85        & 72.58                       & 41.43                    & 51.83             & 41.06             & 54.38             \\
        
        A$^3$lign-DFER\cite{tao20243}          & CLIP-ViT-L/14             & 64.09        & 74.20                       & 41.87                    & 51.77             & 42.07             & 53.24             \\
        OUS\cite{mai2024ous}                  & CLIP-ViT-L/14                  & 60.94                    & 74.10                       & 42.23                    & \textbf{53.30}           & -             & -             \\

        PE-CLIP\cite{saadi2025pe}          & CLIP-ViT-B/16             & 62.82        & 74.04                       & 41.57                    & 51.26             & -             & -             \\
        CLVSR\cite{liang2026clvsr}          & CLIP-ViT-B/16             & \underline{64.33} & 71.58                       & \textbf{43.52} & 50.66             & \underline{42.51} & 52.69             \\





         \textbf{STL (Ours)}       &    CLIP-ViT-B/16                  & 61.85                    & \underline{74.43}                       & 41.10        & 51.71             & 41.81             & \underline{56.15}            \\
        
        
         \textbf{SSM (Ours)}            & CLIP-ViT-B/16        &     \textbf{64.83}        & \textbf{75.37}              &  \underline{43.21}          & \underline{53.28}     & \textbf{43.38}     & \textbf{57.26}   \\        
    \bottomrule
    \end{tabular}

\end{table*}

\begin{table*}[!t]
    \centering
    \caption{Detailed comparison of accuracy across various emotion categories on DFEW.
        The best results are highlighted in bold, and the second-best underlined.}
    \label{tab:specific}
    \begin{adjustbox}{max width=\textwidth}
    \small
    \begin{tabular}{lccccccccc}
        \toprule
        \multirow{2}{*}{Method}                      & \multicolumn{7}{c}{Accuracy of Each Emotion} & \multicolumn{2}{c}{DFEW}                                                                                                                                             \\
        \cmidrule(lr){2-8} \cmidrule(lr){9-10}
                                                     & Happy                                        & Sad                      & Neutral           & Angry             & Surprise          & Disgust           & Fear              & UAR               & WAR               \\
        \hline
        EC-STFL  ~\cite{jiang2020dfew} & 79.18                                        & 49.05                    & 57.85             & 60.98             & 46.15          & 2.76           & 21.51          & 45.35          & 56.51          \\

        Former-DFER  \cite{zhao2021former}     & 84.05                                        & 62.57                    & 67.52             & 70.03             & 56.43             & 3.45              & 31.78             & 53.69             & 65.70             \\
        NR-DFERNet  \cite{li2022nr}            & 88.47                                        & 64.84                    & 70.03             & 75.09             & 61.60             & 0.00              & 19.43             & 54.21             & 68.19             \\
        EST  \cite{liu2023expression}          & 86.87                                        & 66.58                    & 67.18             & 71.84             & 47.53             & 5.52              & 28.49             & 53.43             & 65.85             \\
        IAL  \cite{li2023intensity}            & 87.95                                        & 67.21                    & 70.10             & 76.06             & \underline{62.22}             & 0.00              & 36.44             & 55.71             & 69.24             \\
        M3DFEL  \cite{wang2023rethinking}      & 89.59                                        & 68.38                    & 67.88             & 74.24             & 59.69             & 0.00              & 31.64             & 56.10             & 69.25             \\

        SVFAP \cite{sun2024svfap}             & \textbf{93.13}                                        & 76.98                    & 72.31             & \underline{77.54}             & \textbf{65.42} & 15.17             & 39.25             & 62.83             & 74.27             \\
        MAE-DFER \cite{sun2023mae}             & \underline{92.92}                                        & \underline{77.46}                    & \textbf{74.56}             & 76.94             & 60.99             & \underline{18.62} & \underline{42.35} & \underline{63.41}             & \underline{74.43}             \\

        \textbf{SSM (Ours)}                     & 92.64                               & \textbf{79.83}        & \underline{73.55} & \textbf{79.24} & 61.81    & \textbf{20.95}    & \textbf{45.80}    & \textbf{64.83}    & \textbf{75.37}    \\
        \bottomrule
  \end{tabular}
  \end{adjustbox}
\end{table*}

\subsection{Comparison with the State of the Art}
\subsubsection{Facial Action Unit Detection}
To validate the effectiveness of our method, we compare it with several state-of-the-art methods on BP4D and DISFA, including FMAE-IAT \cite{ning2024representation}, MAE-Face \cite{ma2024facial}, HiVA \cite{li2026hierarchical}, and CausalAffect (+EmotioNet)\footnote{\emph{CausalAffect (+EmotioNet) denotes its best-performing variant using one auxiliary dataset, matching our one-to-one dataset-pairing protocol.}} \cite{hu2025causalaffect}. We select DFEW as the paired dataset because it yields the best DFER performance when jointly learned with AU detection. Table~\ref{tab:bp4d_results} reports the comparison of F1 scores over 12 AUs on BP4D. The results show that SSM performs favorably on multiple AUs. The improvements are particularly notable on AU15, AU17, and AU23. SSM also achieves the highest average F1 score among all compared methods. Table~\ref{tab:disfa_results} presents the results on DISFA. SSM achieves the highest average F1 score over 8 AUs among all compared methods. The most pronounced improvement is observed on AU2. In addition, joint learning with a DFER dataset outperforms single-task learning (SSM \textit{vs.} STL) on both BP4D and DISFA, with gains of +2.3\% on each dataset.  This result shows that fine-grained local AUs can benefit from coarse-grained global expressions. However, the gains are not uniform across AUs. For instance, compared with STL, the F1 score decreases for AU4 on BP4D and for AU6 and AU25 on DISFA, although the average F1 increases on both datasets. DPM is optimized through the joint task objective rather than separately for each AU; therefore, the average improvement does not guarantee gains for every AU. These results show that FE→AU transfer may introduce negative transfer for specific AUs.\footnote{\emph{Detailed AU-wise and failure-case analyses are provided in Sec. H of the supplementary material.}}

\begin{table}[ht]
  \centering
  \scriptsize
  \setlength{\tabcolsep}{4pt}
    \caption{Investigation of the contributions of each component in the SSM framework. Multi-task Baseline: Baseline model in Fig.~\ref{baseline_image} trained jointly on heterogeneous datasets; TSP: Textual Semantic Prototype module; DPM: Dynamic Prior Mapping module. Note that the SSM framework is always built on top of the Baseline model. Metrics on BP4D and DISFA: F1 score. Metrics on DFEW: UAR/WAR.}
  \label{tab:all}
  \resizebox{\columnwidth}{!}{%
    \begin{tabular}{ccc|cc|cc}
      \toprule
      Baseline (MTL) & TSP & DPM & \makecell{BP4D} & \makecell{DFEW} & \makecell{DISFA} & \makecell{DFEW}  \\
      \midrule
       &  &  & 66.2 & 63.98/76.16 & 69.6 & 63.98/76.16   \\
      \cmark &  &  & 67.2 & 65.25/76.97 & 70.4 & 65.93/77.53  \\
      \cmark & \cmark &  & 67.7 & 66.75/77.35 & 70.6 &  66.03/77.78 \\
      \cmark & \cmark & \cmark & \textbf{68.5} &\textbf{68.59}/\textbf{77.88} & \textbf{71.3} & \textbf{66.64}/\textbf{78.09} \\
      \bottomrule
    \end{tabular}%
  }    
\end{table}

\subsubsection{Dynamic Facial Expression Recognition}
To verify the multi-task learning capability of our method, we also conduct experiments on the dynamic facial expression recognition task. We compare our method with the advanced methods on DFEW, FERV39K, and MAFW, which can be divided into three paradigms: supervised learning methods, self-supervised learning methods, and vision--language models. They include IFDD-3DViT \cite{wang2025lifting}, SVFAP \cite{sun2024svfap}, MAE-DFER \cite{sun2023mae}, A$^3$lign-DFER \cite{tao20243}, OUS \cite{mai2024ous}, PE-CLIP \cite{saadi2025pe}, and CLVSR \cite{liang2026clvsr}. Similarly, we select BP4D as the paired dataset because it yields the best AU detection performance when jointly learned with DFER. The results are shown in Table~\ref{tab:dfer}. SSM achieves performance on par with state-of-the-art methods across all three datasets. Moreover, SSM outperforms the single-task model, STL, on all three datasets: DFEW (UAR: +2.98\%, WAR: +0.94\%), FERV39K (UAR: +2.11\%, WAR: +1.57\%), and MAFW (UAR: +1.57\%, WAR: +1.11\%). This result indicates that AUs collected under laboratory conditions can also facilitate in-the-wild expression recognition. Together with the results in Table~\ref{tab:bp4d_results} and Table~\ref{tab:disfa_results}, this finding answers our earlier question. Under dynamic settings, the inherent AU–FE relationship can be further exploited in both directions across heterogeneous datasets. In addition, Table~\ref{tab:specific} reports the recognition accuracy for each of the seven expression categories in DFEW. The results show that the accuracy of the two low-sample classes, fear and disgust, is also improved.

\subsection{Ablation Studies}
\textbf{Key Component Ablation}: To evaluate the effectiveness of each component in SSM, we conduct extensive ablation studies. To avoid the enormous computational cost caused by Cartesian-product-style dataset combinations\footnote{\emph{The experimental results of the Cartesian-product-based dataset combinations under the SSM framework are listed in Table~\ref{tab:dataset_all} in Sec.~\ref{supp-sec:Exhaustive-Results}.}}, and to cover a broader range of data, we perform all ablation experiments on three folds of BP4D and DISFA and one fold of DFEW. We use BP4D+DFEW as one multi-task learning group and DISFA+DFEW as the other. We validate the Baseline model (Baseline), the Textual Semantic Prototype (TSP) module, and the adaptive Dynamic Prior Mapping (DPM) module. Notably, the text encoder is used only when TSP is enabled, and TSP is indispensable for DPM. 

Table~\ref{tab:all} reports the performance of different component combinations. Specifically, the Baseline model alone already yields clear gains, i.e., BP4D (F1 score: +1.0\%) and DFEW (UAR: +1.27\%, WAR: +0.81\%), as well as DISFA (F1 score: +0.8\%) and DFEW (UAR: +1.95\%, WAR: +1.37\%). DPM transfers knowledge through a textual medium and dynamically adjusts during training, making cross-task transfer effective. It brings noticeable improvements on BP4D (F1 score: +0.8\%) and DFEW (UAR: +1.84\%, WAR: +0.53\%), as well as on DISFA (F1 score: +0.7\%) and DFEW (UAR: +0.61\%, WAR: +0.31\%). TSP directly supports the DPM module. Compared with traditional one-hot labels, TSP provides a more unified deep semantic space for the two tasks. For instance, the two AU labels ``brow lowerer'' and ``brow raiser'' are completely unrelated in a discrete one-hot label space, whereas in a textual semantic space they are pulled closer because they share the word ``brow''. Obvious performance increase can be seen on BP4D (F1 score: +0.5\%) and DFEW (UAR: +1.50\%, WAR: +0.38\%), as well as on DISFA (F1 score: +0.2\%) and DFEW (UAR: +0.10\%, WAR: +0.25\%), brought by TSP. Together with the gains achieved by the Baseline, the further improvements brought by TSP and DPM indicate that semantic transfer complements shared visual learning under heterogeneous AU--FE supervision.

\begin{table}[ht]
  \centering
  \scriptsize
  \setlength{\tabcolsep}{4pt} 
\caption{How to construct an effective DPM module. We further probe its working mechanism by constructing the DPM in different ways. R init.: random initialization; P init.: prior initialization; Dual: whether bidirectional independent learning is performed. Metrics on BP4D and DISFA: F1 score. Metrics on DFEW: UAR/WAR.}
  \label{tab:DPM1}
  \resizebox{\columnwidth}{!}{%
    \begin{tabular}{ccc|cc|cc}
      \toprule
      R init. & P init. & Dual & \makecell{BP4D} & \makecell{DFEW} & \makecell{DISFA} & \makecell{DFEW}  \\
      \midrule
      \cmark &  &  & 67.0 & 65.35/77.35 & 69.6  & 64.52/76.88    \\
      \cmark &  & \cmark & 67.2 & 65.75/77.53 & 70.2 & 65.20/76.84   \\
        & \cmark &  & 68.1 & 67.00/77.01 & 70.9 & 66.37/77.23  \\
        & \cmark & \cmark & \textbf{68.5} &\textbf{68.59}/\textbf{77.88} & \textbf{71.3} & \textbf{66.64}/\textbf{78.09} \\
      \bottomrule
    \end{tabular}%
  }
\end{table}

\textbf{Dynamic Prior Mapping (DPM)}: To better understand the working mechanism of DPM, we conduct a deeper ablation analysis, as shown in Table~\ref{tab:DPM1}. The experimental results indicate that prior-guided DPM improves task performance (R init. \textit{vs.} P init.). Moreover, our bidirectional learning strategy differs from a simple matrix transpose. It allows the two mapping directions to adapt independently to heterogeneous data distributions and improves model performance (w/ Dual \textit{vs.} w/o Dual). Under the prior-initialized setting, the bidirectional learning strategy brings clear gains, i.e., BP4D (F1 score: +0.4\%) and DFEW (UAR: +1.59\%, WAR: +0.87\%), as well as DISFA (F1 score: +0.4\%) and DFEW (UAR: +0.27\%, WAR: +0.86\%). This result is consistent with prior findings \cite{li2023compound}.

\begin{table}[ht]
  \centering
  \scriptsize
  \setlength{\tabcolsep}{6pt} 
\caption{Effectiveness of the DPM module. We compare the full DPM with Linear, MLP, DPM (Random, Frozen), and DPM (Prior, Frozen). ``Frozen'' indicates that the DPM weight matrices are fixed during training. Metrics on BP4D and DISFA: F1 score. Metrics on DFEW: UAR/WAR.}
  \label{tab:DPM2}
  \resizebox{\columnwidth}{!}{%
    \begin{tabular}{c|cc|cc}
      \toprule
      Setting & \makecell{BP4D} & \makecell{DFEW} & \makecell{DISFA} & \makecell{DFEW}  \\
      \midrule
      Linear & 66.8 &64.79/76.87 &69.7  &64.77/76.88    \\
      MLP & 67.5 &65.26/77.23 &70.6  &65.89/77.18    \\
      DPM (Random, Frozen) & 66.9 &65.12/76.61 &69.7 &64.73/76.84   \\
      DPM (Prior, Frozen) & 67.3 &66.62/77.05 &70.2 &65.47/77.53   \\
      DPM & \textbf{68.5} &\textbf{68.59}/\textbf{77.88} & \textbf{71.3} & \textbf{66.64}/\textbf{78.09} \\    
      \bottomrule
    \end{tabular}%
  }
\end{table}


In addition, we conduct replacement-style ablation studies for DPM. We compare three groups of results, namely Linear, MLP, and several DPM variants, as reported in Table~\ref{tab:DPM2}. The results show that prior-guided DPM is clearly superior to the other two groups. They also show that MLP performs better than Linear. This further confirms that simple mappings are insufficient to capture complex correspondences and therefore degrade model performance. Furthermore, to examine the role of data-driven adaptation, we set up a control comparison between learnable and non-learnable DPM, namely, (Prior, Frozen) \textit{vs.} DPM in Table~\ref{tab:DPM2}. The non-learnable DPM is fixed after FACS-based initialization and cannot be dynamically adjusted according to data characteristics. As a result, it leads to performance drops on BP4D (F1 score: -1.2\%) and DFEW (UAR: -1.97\%, WAR: -0.83\%), as well as on DISFA (F1 score: -1.1\%) and DFEW (UAR: -1.17\%, WAR: -0.56\%).\footnote{\emph{Further controlled analyses of auxiliary-task data, the fixed FACS-informed prior, and dynamic adjustment are provided in Sec.~G of the supplementary material.}} 

\textbf{Textual Semantic Prototype (TSP)}: 
It is necessary to investigate the impact of different text descriptions. 
We have tried three types of text descriptions, i.e., Compound (e.g., ``cheek raiser, lip corner puller''), Standalone (e.g., ``a facial expression of happiness''), and Words (e.g., ``happiness'').\footnote{\emph{Specifically, the detailed compound descriptions are listed in Table S4 in Sec. D of the supplementary material.}} Table~\ref{tab:D} shows the effects of different label description forms. The results indicate that Compound descriptions improve task performance the most. 

\begin{table}[ht]
  \centering
  \scriptsize
  \setlength{\tabcolsep}{6pt} 
  \caption{How to construct an effective textual semantic prototype module (TSP). Since DPM is always based on TSP, the effect of knowledge transfer also indirectly depends on the type of text. We use different textual descriptions to investigate their impact on the model. Words: word descriptions; Standalone: standalone descriptions; Compound: composite descriptions. BP4D and DISFA: F1 score. DFEW: UAR/WAR.}
  \label{tab:D}
  \resizebox{\columnwidth}{!}{%
    \begin{tabular}{c|cc|cc}
      \toprule
      Setting & \makecell{BP4D} & \makecell{DFEW} & \makecell{DISFA} & \makecell{DFEW}  \\
      \midrule
      Words & 68.2 &64.56/77.87 &68.1  &\textbf{67.24}/77.40    \\
      Standalone & 68.1 &65.33/77.31 &69.7  &65.60/77.53    \\
      Compound & \textbf{68.5} &\textbf{68.59}/\textbf{77.88} & \textbf{71.3} & 66.64/\textbf{78.09} \\ 
      \bottomrule
    \end{tabular}%
  } 
\end{table}

Since CLIP's text encoder is always kept frozen, the learnable tokens become the medium through which DPM connects the text and vision branches. We further explore the number of such tokens, as shown in Table~\ref{coop}. Using either too many or too few tokens leads to performance degradation. 

\begin{table}[ht]
  \centering
  \scriptsize
  \setlength{\tabcolsep}{6pt} 
  \caption{Analysis of the number of learnable tokens in text descriptions. BP4D and DISFA: F1 score. DFEW: UAR/WAR.}
  \label{coop}
  \resizebox{\columnwidth}{!}{%
    \begin{tabular}{c|cc|cc}
      \toprule
       Prompt Count & \makecell{BP4D} & \makecell{DFEW} & \makecell{DISFA} & \makecell{DFEW}  \\
      \midrule
      0 & 67.9 &63.70/77.23 &70.2  &63.84/76.67    \\
      4 & 66.8 &67.49/77.05 &68.6  &64.95/77.06    \\
      8 & \textbf{68.5} &\textbf{68.59}/\textbf{77.88} & \textbf{71.3} & \textbf{66.64}/\textbf{78.09} \\
      12 & 67.0 &66.15/77.48 &69.6  &66.50/77.19    \\
      16 & 66.6 &65.96/77.53 &69.5  &64.82/77.05    \\
      \bottomrule
    \end{tabular}%
  } 
\end{table}

\begin{figure}[t]  
    \centering
    \includegraphics[width=\columnwidth]{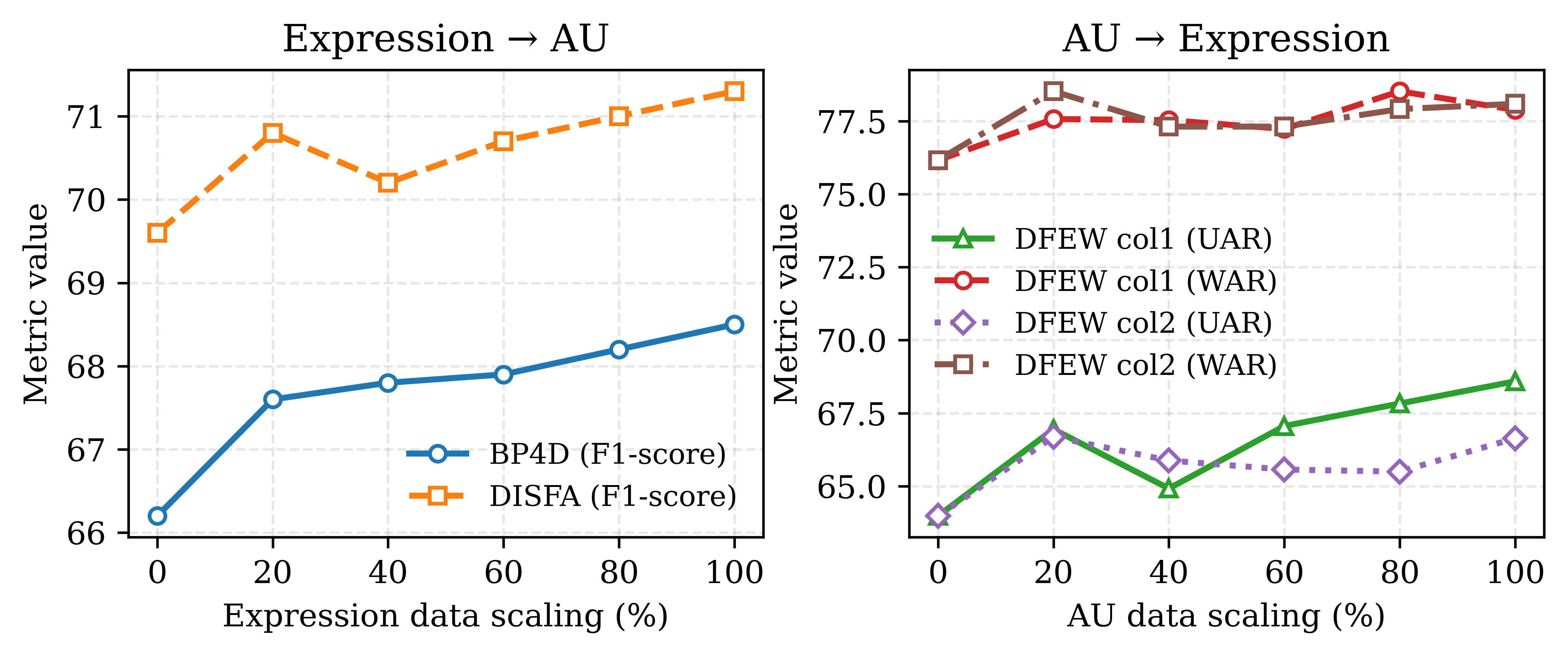}
    \caption{The analysis of data scale, where 0\% data scale indicates single-task training. In the left panel (Expression $\rightarrow$ AU), BP4D and DISFA denote joint learning with DFEW. In the right panel (AU $\rightarrow$ Expression), DFEW col1 denotes joint learning with BP4D, and DFEW col2 denotes joint learning with DISFA.}
    \label{data_scale}
\end{figure}

\textbf{Data Scaling Study}: Finally, we quantitatively study how the data scale of one task affects the other under joint learning. For the AU detection task, we use 100\% of the AU data and progressively use 20\%, 40\%, 60\%, 80\%, and 100\% of the FE data to investigate the effect of FE$\to$AU. The same protocol is applied to the DFER task. The upper part of Fig.~\ref{data_scale} presents the quantitative analysis of FE$\to$AU, while the lower part presents the quantitative analysis of AU$\to$FE.\footnote{\emph{The specific metrics are provided in Table S1 and Table S2 in Sec. A of the supplementary material.}} We observe that positive gains already appear when only 20\% of the paired-task data are used, and these gains are generally maintained as the data scale increases. This effect becomes more pronounced as the data scale increases. This result suggests that the gains brought by SSM cannot be explained solely by an increase in the amount of paired-task data.

\begin{figure*}[ht!]
  \centering
\includegraphics[width=\textwidth,height=\textheight,keepaspectratio]{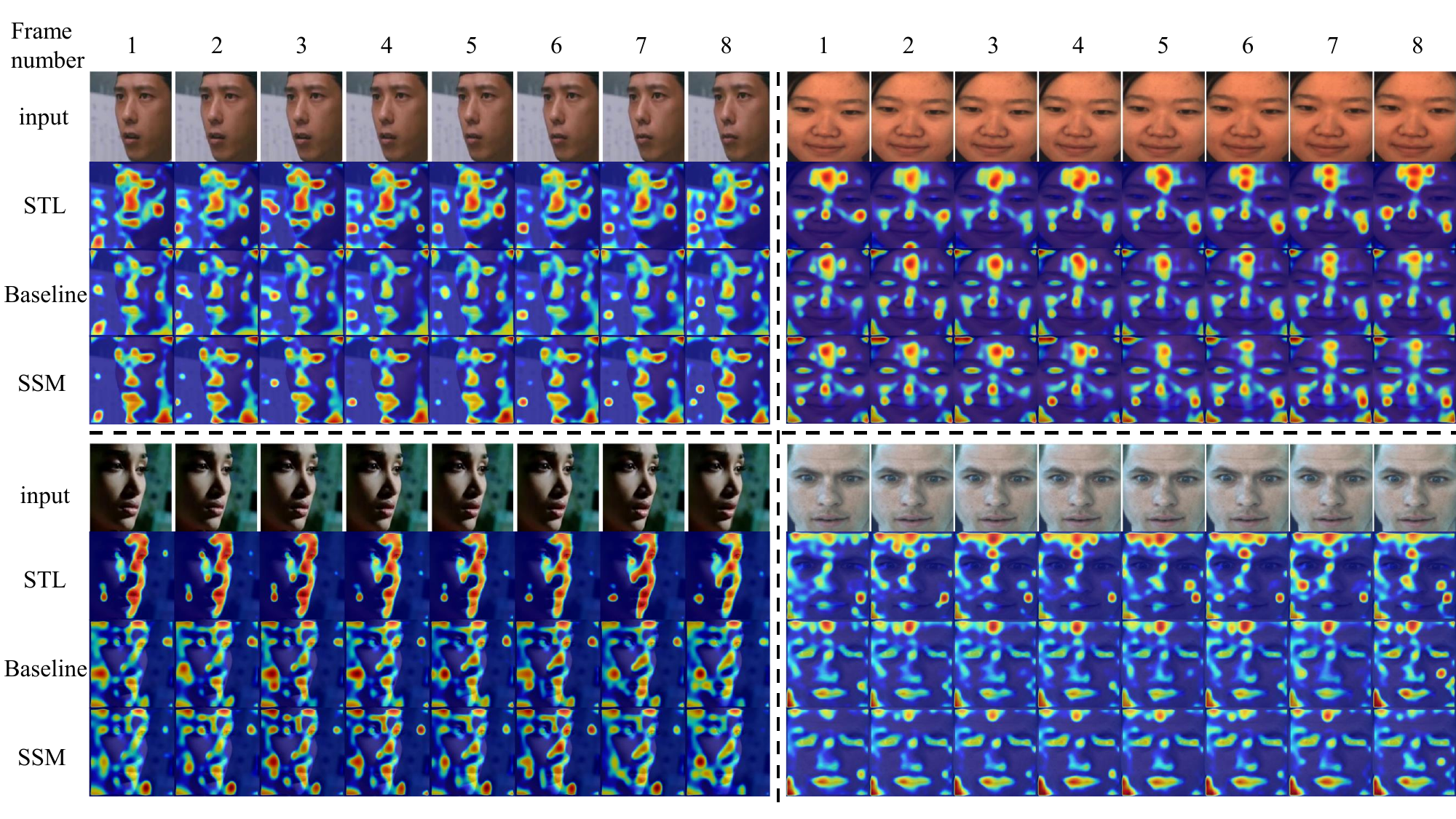}
\caption{Visualization of the input facial frame sequences. The left side shows samples from the DFER task. The right side shows samples from the AU detection task. Each subfigure presents an overlay of the original face frame and the attention heatmap. Warmer colors indicate regions with higher responses. The attention maps are computed from the multi-head self-attention of the CLIP-ViT-B/16 visual encoder using the attention rollout method \cite{abnar2020quantifying}. From top to bottom, the attention overlays correspond to STL, Baseline, and SSM, respectively. Compared with STL, Baseline attends to more facial regions. Compared with Baseline, SSM shows denser and more spatially organized responses across frames.}
  \label{attention}
\end{figure*}

\begin{figure*}[ht!]
  \centering
\includegraphics[width=\textwidth,height=\textheight,keepaspectratio]{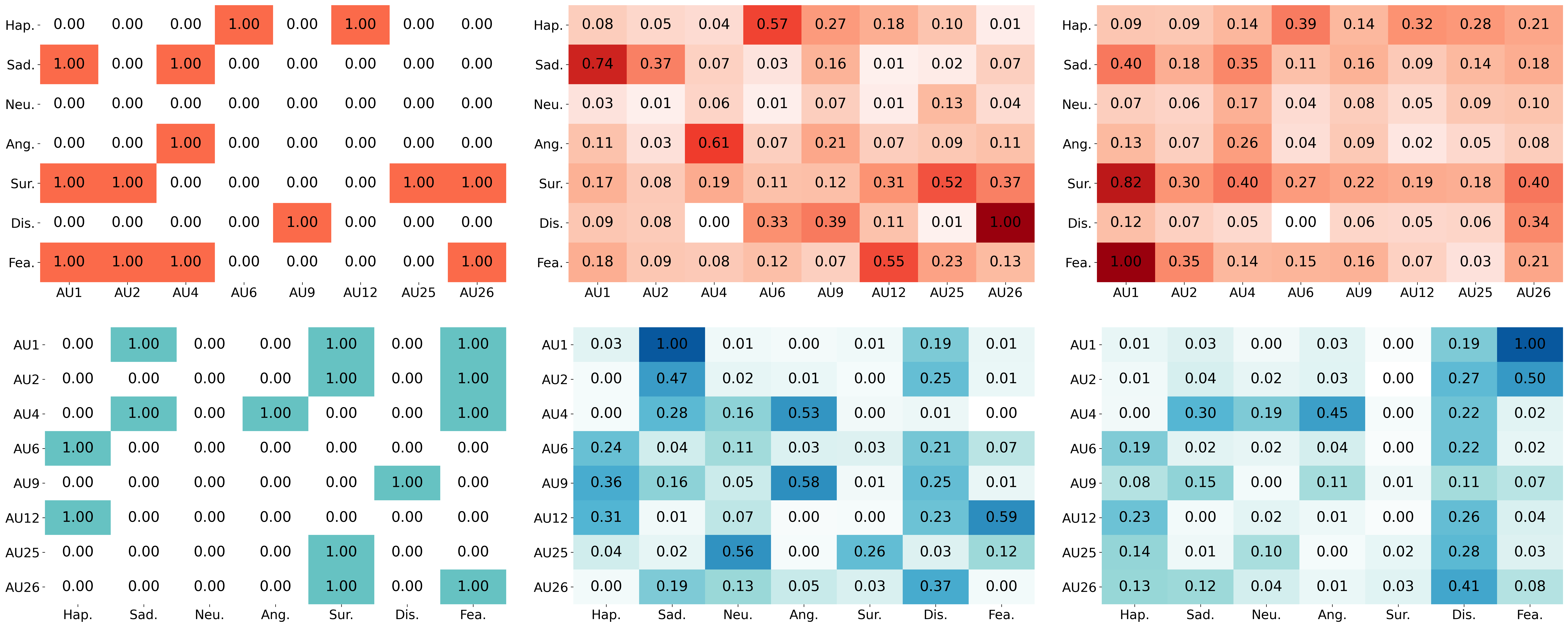}
\caption{The upper part shows the weight matrix activation maps from AUs to facial expressions on DISFA and DFEW (fd5). From left to right, the three matrices correspond to the initial matrix, the matrix learned from random initialization, and the matrix learned from prior-based initialization. The lower part shows the corresponding weight matrices from facial expressions to AUs.}
  \label{weight_matrix}
\end{figure*}


\subsection{Cross-Domain Evaluation}
We evaluate our model in a zero-shot setting. Specifically, we train on the combination of BP4D and DFEW. We then conduct zero-shot testing on the combination of DISFA and FERV39K. For testing on DISFA, we index the output distribution of BP4D to match the shared labels in DISFA. This results in five AUs in total, namely AU1, AU2, AU4, AU6, and AU12, and we report their average F1 score. For testing on FERV39K, its label distribution is consistent with DFEW, so we directly conduct the evaluation.

\begin{table}[ht]
  \centering
  \scriptsize
  \setlength{\tabcolsep}{6pt}
  \caption{Cross-dataset testing. We train on the BP4D+DFEW combination and perform zero-shot testing on the DISFA+FERV39K combination. We compare STL, Baseline, and SSM. For the DISFA test, we select the five AUs shared by BP4D and DISFA and report their average F1 score. BP4D and DISFA: F1 score. DFEW and FERV39K: UAR/WAR.}
  \label{cross}
  \resizebox{\columnwidth}{!}{%
    \begin{tabular}{c|cc|cc}
      \toprule
      & \multicolumn{2}{c|}{Train} & \multicolumn{2}{c}{Test} \\
      \cmidrule(lr){2-3} \cmidrule(lr){4-5}
       & BP4D & DFEW & DISFA & FERV39K \\
      \midrule
      STL & 66.2 & 63.98/76.16 & 46.5 & 29.91/39.17 \\
      Baseline & 67.2 & 65.25/76.97 & 59.4 & 31.55/41.98 \\
      SSM & \textbf{68.5} &\textbf{68.59}/\textbf{77.88}  & \textbf{67.1} & \textbf{32.10}/\textbf{43.52} \\
      \bottomrule
    \end{tabular}%
  }
\end{table}

The results are shown on the right side of Table~\ref{cross}. We compare the single-task model, our Baseline model, and the final SSM framework. Cross-domain zero-shot testing is challenging. The performance drop is much larger on DFER than on AU detection, which is expected because BP4D and DISFA are relatively closer in domain characteristics and label space, whereas DFEW and FERV39K differ more substantially in both data domain and annotation protocol \cite{chen2024static,chen2025static}. Nevertheless, the results exhibit a consistent trend. The SSM framework clearly outperforms our Baseline model, i.e., DISFA (F1 score: +7.7\%) and FERV39K (UAR: +0.55\%, WAR: +1.54\%). Moreover, our Baseline model also clearly outperforms the single-task model, i.e., DISFA (F1 score: +12.9\%) and FERV39K (UAR: +1.64\%, WAR: +2.81\%). These results show better cross-dataset transfer under joint learning, with SSM giving the strongest results among the three settings.\footnote{\emph{Further cross-dataset representation and AU--FE label-space analyses are provided in Sec.~J of the supplementary material.}}

\subsection{Exhaustive Results over Different Dataset Pairings}
\label{supp-sec:Exhaustive-Results}


Table~\ref{tab:dataset_all} provides a more comprehensive summary of joint-learning results across different dataset and fold combinations. We adopt an enumeration strategy based on the Cartesian product of fold pairings. This reduces the dependence on specific pairing choices. Across all relevant fold pairings in Table~\ref{tab:dataset_all}, the mean results are 67.53 on BP4D, 70.67 on DISFA, 42.20/52.69 on FERV39K, 63.27/75.23 on DFEW, and 42.97/56.93 on MAFW. The corresponding STL averages over the official folds are 66.17, 69.63, 41.10/51.71, 61.85/74.43, and 41.81/56.15, respectively, indicating that the overall improvement trend of SSM over STL remains consistent when different fold pairings are considered.

\begin{table*}[t]
\centering
\caption{Summary of results across different dataset combinations. For datasets with varying numbers of folds, we adopt a Cartesian-product-based combination strategy for exhaustive training, thereby reducing the dependence on specific dataset pairings. Metrics on BP4D and DISFA: F1 score. Metrics on FERV39K, DFEW, and MAFW: UAR/WAR.}
\label{tab:dataset_all}
\begin{adjustbox}{max width=\textwidth}
\small
\begin{tabular}{lc|ccccc|ccccc}
\toprule
 & FERV39K & DFEW\_fd1 & DFEW\_fd2 & DFEW\_fd3 & DFEW\_fd4 & DFEW\_fd5 & MAFW\_fd1 & MAFW\_fd2 & MAFW\_fd3 & MAFW\_fd4 & MAFW\_fd5 \\
\midrule
BP4D\_fd1 & \makecell[c]{64.3\\42.63/53.46} & \makecell[c]{67.0\\61.38/75.98} & \makecell[c]{66.9\\60.83/72.39} & \makecell[c]{66.4\\62.37/74.27} & \makecell[c]{66.4\\64.10/75.34} & \makecell[c]{67.5\\68.59/77.88} & \makecell[c]{68.2\\36.8/49.40} & \makecell[c]{67.8\\42.14/54.48} & \makecell[c]{67.7\\45.47/58.84} & \makecell[c]{67.7\\47.50/61.30} & \makecell[c]{67.6\\44.06/59.48} \\

BP4D\_fd2 & \makecell[c]{67.8\\42.62/53.19} & \makecell[c]{70.3\\61.21/76.03} & \makecell[c]{70.5\\60.85/72.39} & \makecell[c]{70.0\\61.44/74.06} & \makecell[c]{69.7\\62.71/75.26} & \makecell[c]{70.4\\67.11/77.79} & \makecell[c]{69.6\\37.18/50.05} & \makecell[c]{68.6\\41.64/54.48} & \makecell[c]{68.6\\45.58/59.33} & \makecell[c]{69.0\\47.76/61.08} & \makecell[c]{69.0\\44.57/59.87} \\

BP4D\_fd3 & \makecell[c]{66.9\\43.21/53.28} & \makecell[c]{68.1\\64.97/76.32} & \makecell[c]{68.1\\64.11/73.03} & \makecell[c]{67.9\\61.14/74.27} & \makecell[c]{67.8\\62.43/75.17} & \makecell[c]{67.6\\67.24/77.62} & \makecell[c]{63.9\\36.78/49.95} & \makecell[c]{65.2\\42.25/55.35} & \makecell[c]{63.6\\45.19/58.84} & \makecell[c]{64.5\\47.34/61.68} & \makecell[c]{64.0\\44.30/59.54} \\

\midrule

DISFA\_fd1 & \makecell[c]{73.9\\42.27/52.53} & \makecell[c]{74.6\\62.32/75.60} & \makecell[c]{74.4\\63.38/73.33} & \makecell[c]{74.8\\62.89/74.96} & \makecell[c]{74.0\\63.28/74.79} & \makecell[c]{74.8\\66.64/78.09} & \makecell[c]{72.0\\36.35/49.07} & \makecell[c]{73.5\\39.73/54.64} & \makecell[c]{73.6\\46.17/59.84} & \makecell[c]{73.5\\44.61/60.60} & \makecell[c]{73.9\\44.84/60.38} \\

DISFA\_fd2 & \makecell[c]{70.7\\41.18/51.87} & \makecell[c]{71.9\\62.24/75.73} & \makecell[c]{71.5\\62.88/73.12} & \makecell[c]{72.5\\60.69/74.40} & \makecell[c]{71.1\\63.28/75.52} & \makecell[c]{71.4\\66.11/77.84} & \makecell[c]{73.6\\36.09/48.63} & \makecell[c]{72.4\\42.28/54.70} & \makecell[c]{71.9\\45.97/60.33} & \makecell[c]{72.7\\44.78/60.44} & \makecell[c]{73.9\\44.69/60.27} \\

DISFA\_fd3 & \makecell[c]{67.1\\41.29/51.79} & \makecell[c]{67.1\\62.27/75.56} & \makecell[c]{67.6\\62.03/73.08} & \makecell[c]{68.5\\62.12/74.57} & \makecell[c]{66.8\\60.93/74.87} & \makecell[c]{67.7\\66.55/77.75} & \makecell[c]{63.9\\36.37/49.02} & \makecell[c]{63.8\\40.60/54.43} & \makecell[c]{65.2\\46.08/59.73} & \makecell[c]{63.8\\46.43/60.87} & \makecell[c]{63.9\\45.49/61.26} \\
\bottomrule
\end{tabular}
\end{adjustbox}
\end{table*}

\subsection{Visualization}
\subsubsection{Attention Visualization}
Fig.~\ref{attention} presents the attention heatmaps of the single-task model, the Baseline model, and SSM on image samples from several datasets. From STL $\to$ Baseline $\to$ SSM, the attention pattern evolves from ``few and local (coarse-grained)'' to ``more and structured (fine-grained).'' Specifically, STL mainly focuses on a few salient regions, such as the mouth or local eyebrow--eye regions. This indicates a reliance on a single discriminative cue. In DFER, such attention may overlook the coordinated dynamics of expression-related muscle groups. In AU detection, it may also miss auxiliary regions that co-occur with the target AU. The Baseline introduces cross-task supervision. It encourages the model to shift from single-point evidence to multi-region evidence fusion. As a result, the attention coverage expands, although it is often broader and more scattered. Building on the Baseline, SSM further improves the cross-task semantic transfer mechanism by leveraging TSP and DPM, leading to stronger and more coordinated attention responses. Unlike the Baseline, which mainly broadens the attended regions, SSM better emphasizes informative facial cues while preserving multi-region attention. This results in a more refined attention pattern and facilitates knowledge transfer between FEs and AUs. Importantly, this ``dispersion'' does not indicate ineffective diffusion. Instead, it reflects a shift from dependence on single-point features to joint modeling of multiple muscle groups. This response pattern is more consistent with the local muscle semantics of AUs and the global configurational characteristics of DFER. It is also consistent with the trend of quantitative performance improvement.

\subsubsection{Weight Matrix Visualization}
We visualize the bidirectional weight contribution matrices between AUs and expressions on the combined DISFA and DFEW datasets, as shown in Fig.~\ref{weight_matrix}. The prior-initialized matrices are not identical to the initially defined weights. Moreover, the two matrices learned bidirectionally are not transposes of each other. This indicates that SSM has already learned to adapt to actual heterogeneous data conditions. Notably, because the weights can be adjusted freely, even randomly initialized matrices can eventually learn some correct weights. This is sufficient to demonstrate the strong capability of SSM.\footnote{\emph{Additionally, we visualize the weight matrices for each dataset combination. Details are illustrated in Figs. S1 and S2 in Sec. C of the supplementary material.}}

\subsubsection{Analysis of the Initial Weighting Factor}
We further analyze the influence of the initial weighting factor in Fig.~\ref{hyperparam}.\footnote{\emph{The specific metrics are listed in Table S3 in Sec. B of the supplementary material.}} Specifically, the coefficients $\alpha$ and $\beta$ in the DPM module are varied over \{0.01, 0.05, 0.1, 0.5, 1.0\}. The results show that the performance trends of both tasks remain stable across different settings.
\begin{figure}[!h]
  \centering
  \includegraphics[width=\linewidth]{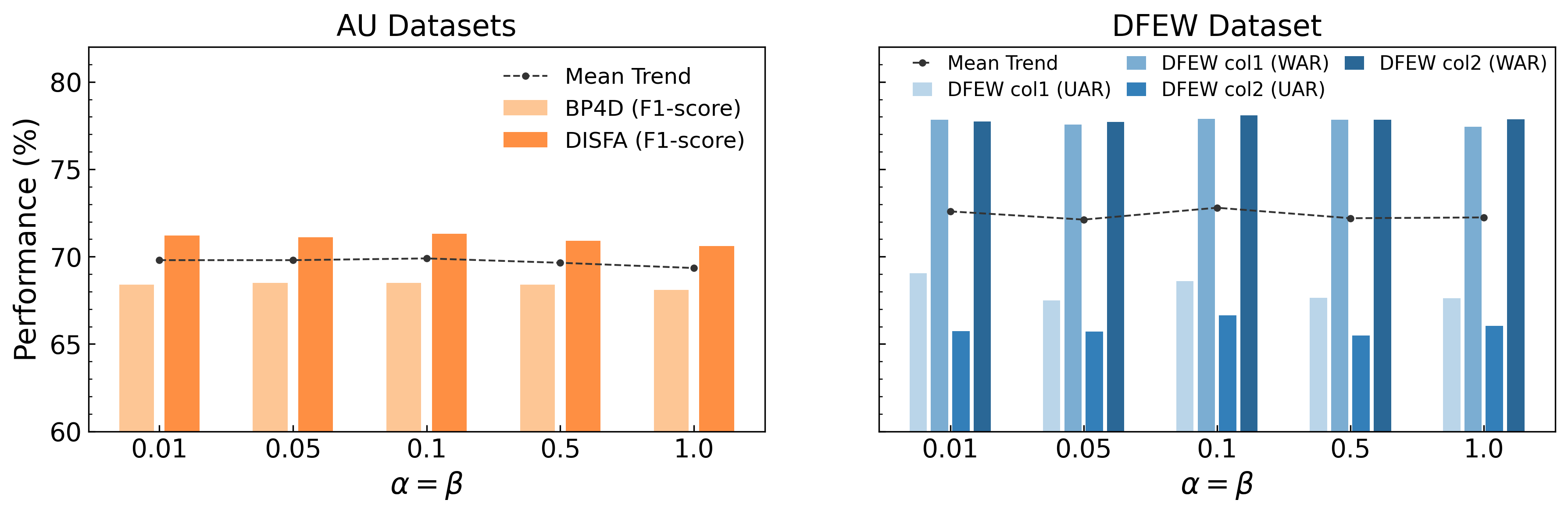}
  \caption{Sensitivity analysis of the initial values of  $\alpha$ and $\beta$ in Eqn. \ref{eq:alpha} and Eqn. \ref{eq:beta}. Left: performance on AU datasets (BP4D and DISFA), measured by F1 score. Right: performance on DFEW, measured by UAR and WAR. The dashed lines indicate the average performance across datasets. The results show that SSM maintains stable performance over a wide range of $\alpha=\beta$. The performance shows a mild peak around 0.1. DFEW col1: jointly learned with BP4D. DFEW col2: jointly learned with DISFA.}
  \label{hyperparam}
\end{figure}


\section{Discussion}

\textbf{Cognitive Perspective.}
The experimental results suggest that AU detection and facial expression recognition can provide complementary information under heterogeneous joint learning. In our setting, the two tasks improve together rather than only in one direction. This indicates that semantic relations between global expressions and local facial actions can still be useful even when the datasets are collected under different conditions and have unaligned annotations. From this perspective, the main value of SSM is that it offers a practical way to connect the two tasks through semantic-level interactions instead of requiring aligned labels or shared dataset design.

\textbf{Model Perspective.}
From a modeling perspective, the final gains come from the combination of several components rather than from a single design choice. The ablation studies show that the baseline joint-learning setting already brings improvements, while the full framework gives more consistent gains. The results further support the role of semantic descriptions, adaptive prior mapping, and bidirectional optimization in the final model behavior. In addition, the comparison with simpler mapping variants suggests that the proposed semantic mapping design is more suitable for this heterogeneous setting than direct or fixed alternatives.

\textbf{Limitations.}
Despite these results, the proposed framework still has several limitations. First, the method remains sensitive to how the text semantics are constructed, because different text forms and prompt settings lead to different results. Second, although the framework improves cross-dataset transfer, the zero-shot setting is still challenging, which means that domain differences are not fully resolved. Third, the current design models cross-task relations mainly at the semantic and dataset levels. It does not explicitly model finer sample-level correspondences or more advanced multimodal interactions. These issues should be studied further in future work.
\section{Conclusion}

In this work, we study joint learning of facial action units (AUs) and facial expressions (FEs) from heterogeneous datasets with unaligned annotations and domain differences. To address this setting, we propose the Structured Semantic Mapping (SSM) framework, which builds semantic-level interactions between the two tasks through textual semantic prototypes and dynamic prior mapping. Experimental results show that the proposed framework improves both AU detection and dynamic facial expression recognition under joint learning. The ablation results further indicate that the performance gains come from the combined effect of semantic descriptions, adaptive mapping, and bidirectional optimization. In addition, the cross-dataset results suggest that the proposed framework has better transfer ability than the compared baselines in the zero-shot setting. Overall, this work shows that heterogeneous facial behavior datasets with non-overlapping annotations can still be used jointly through semantic-level mapping. In future work, we will further study finer-grained sample-level interactions and extend the framework to more complex temporal and multimodal settings.

\bibliographystyle{IEEEtran}
\bibliography{main}

\setcounter{figure}{0}
\setcounter{table}{0}
\setcounter{equation}{0}
\setcounter{section}{0}

\renewcommand{\thefigure}{S\arabic{figure}}
\renewcommand{\thetable}{S\arabic{table}}
\renewcommand{\theequation}{S\arabic{equation}}
\begin{bibunit}[IEEEtran]
\makeatletter
\renewcommand{\@biblabel}[1]{[S#1]}
\renewcommand{\bibcite}[2]{\global\@namedef{b@#1}{S#2}}
\makeatother

\clearpage
\setcounter{page}{1}

\section*{Appendix}
\subsection{Data Scaling Study}
\label{supp-sec:data-volume}

Tables~\ref{tab:FE Data Scaling} and \ref{tab:AU Data Scaling} investigate how the scale of auxiliary-task data affects the target task from two opposite directions. The former corresponds to the Expression \(\rightarrow\) AU setting, whereas the latter corresponds to the AU \(\rightarrow\) Expression setting. As the amount of expression data increases, AU detection performance on both BP4D and DISFA generally improves, although it does not increase monotonically at every data scale. The paired DFEW branch also shows consistent gains. In the reverse setting, expression recognition also benefits from the gradual introduction of AU data. However, the best performance does not strictly coincide with the largest AU data scale. This phenomenon suggests that the gains of SSM cannot be simply attributed to scaling up the auxiliary-task data. Instead, they are more consistent with the complementary effects of cross-task semantic transfer under joint supervision. Overall, coarse-grained expression semantics provide complementary constraints for local AU modeling. In turn, fine-grained AU information enhances expression discrimination.

\begin{table}[ht]
  \centering
  \scriptsize
  \setlength{\tabcolsep}{6pt} 
  \caption{Study of data scaling (Expression → AU). We examine how expression datasets of different scales affect AU detection. This analysis shows that the improvement of the framework on AU detection is not merely caused by the increased scale of expression data. BP4D and DISFA: F1 score. DFEW: UAR/WAR.}
  \label{tab:FE Data Scaling}
  \resizebox{\columnwidth}{!}{%
    \begin{tabular}{c|cc|cc}
      \toprule
      FE data scaling & \makecell{BP4D} & \makecell{DFEW} & \makecell{DISFA} & \makecell{DFEW}  \\
      \midrule
      0\% & 66.2 &-  &69.6  &-    \\
      20\% & 67.6 &57.79/71.40  &70.8  &59.05/71.36    \\
      40\% & 67.8 &60.76/73.50  &70.2  &62.71/74.19    \\
      60\% & 67.9 &63.42/75.39  &70.7 &64.64/75.39  \\
      80\% & 68.2 &65.23/77.31  & 71.0 & 65.12/76.33 \\    
      100\% & \textbf{68.5} &\textbf{68.59}/\textbf{77.88} & \textbf{71.3} & \textbf{66.64}/\textbf{78.09} \\ 
      \bottomrule
    \end{tabular}%
  }
\end{table} 

\begin{table}[ht]
  \centering
  \scriptsize
  \setlength{\tabcolsep}{6pt} 
  \caption{Study of data scaling (AU → Expression). We examine how AU datasets of different scales affect expression recognition. This analysis shows that the framework's performance gains on expression recognition are not merely caused by the increased scale of AU data. BP4D and DISFA: F1 score. DFEW: UAR/WAR.}
  \label{tab:AU Data Scaling}
  \resizebox{\columnwidth}{!}{%
    \begin{tabular}{c|cc|cc}
      \toprule
      AU data scaling & \makecell{BP4D} & \makecell{DFEW} & \makecell{DISFA} & \makecell{DFEW}  \\
      \midrule
      0\% & - &63.98/76.16  &-  &63.98/76.16    \\
      20\% & 67.4 &66.95/77.57  &70.5  &66.69/\textbf{78.52}    \\
      40\% & 68.3 &64.92/77.53  &70.8  &65.88/77.31    \\
      60\% & 67.9 &67.06/77.23  &70.5 &\textbf{67.55}/77.31 \\
      80\% & 68.0 &67.83/\textbf{78.52}  & 71.1 & 65.50/77.91   \\    
      100\% & \textbf{68.5} &\textbf{68.59}/77.88 & \textbf{71.3} & 66.64/78.09 \\ 
      \bottomrule
    \end{tabular}%
  }
\end{table} 

\subsection{Analysis of the Initial Weighting Factor}
\label{supp-sec:Hyperparameter}

Table~\ref{tab:hy} further analyzes the influence of the initial values of  $\alpha$ and $\beta$. These two coefficients control the injection strength of cross-task mapped textual semantics in the residual update. They therefore determine the fusion ratio between the original task semantics and the transferred semantics. The results show only limited performance variation on BP4D, DISFA, and DFEW over a relatively wide value range. This indicates that DPM has favorable robustness. Overall, the most balanced performance is achieved around \(\alpha=\beta=0.1\). This suggests that moderate semantic injection better preserves the discriminability of the original textual representations while still incorporating complementary information from the other task. If the weights are too small, the mapped semantics cannot be fully exploited. If they are too large, the stability of the task-specific semantic representations may be weakened.

\begin{table}[ht]
  \centering
  \scriptsize
  \setlength{\tabcolsep}{6pt} 
  \caption{Analysis of the initial values of weighting factors, \(\alpha\) and \(\beta\), which directly regulate the strength of the DPM module. BP4D and DISFA: F1 score. DFEW: UAR/WAR.}
  \label{tab:hy}
  \resizebox{\columnwidth}{!}{%
    \begin{tabular}{c|cc|cc}
      \toprule
      \(\alpha\), \(\beta\) & \makecell{BP4D} & \makecell{DFEW} & \makecell{DISFA} & \makecell{DFEW}  \\
      \midrule
      0.01 & 68.4 &\textbf{69.05}/77.83 & 71.2 &65.74/77.74    \\
      0.05 & 68.5 &67.49/77.57 &71.1&65.72/77.70    \\
      0.1 & \textbf{68.5} &68.59/\textbf{77.88} & \textbf{71.3} & \textbf{66.64}/\textbf{78.09} \\
      0.5 & 68.4 &67.64/77.83 &70.9 &65.50/77.83   \\
      1.0 & 68.1 &67.63/77.44 & 70.6 & 66.04/77.87 \\    
      \bottomrule
    \end{tabular}%
  }
\end{table}

\subsection{Visualization of Bidirectional Weight Matrices}
\label{supp-sec:Visualization}

\begin{figure*}[ht!]
  \centering
\includegraphics[width=\textwidth,height=\textheight,keepaspectratio]{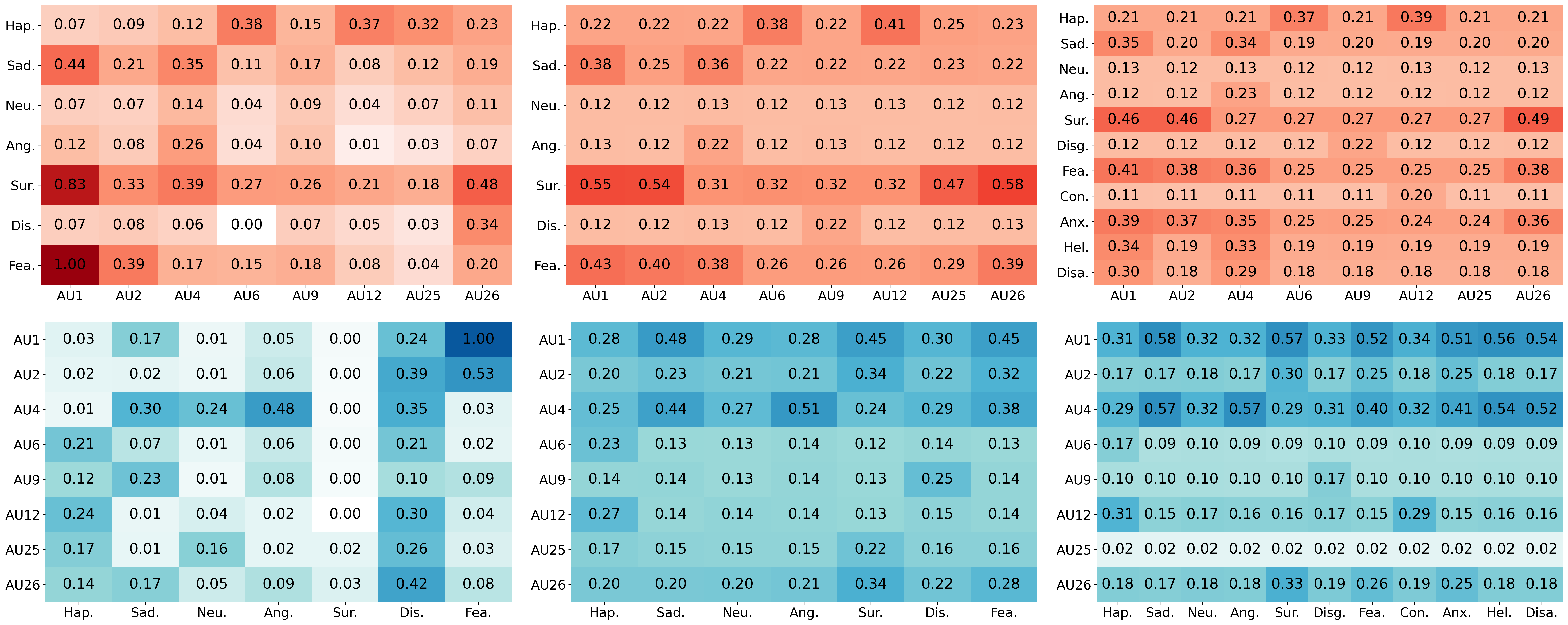}
\caption{The weight matrix activation maps, from left to right, represent joint learning of DISFA with DFEW, FERV39K, and MAFW, respectively. The first row shows the weights from AUs to expressions, and the second row shows the weights from expressions to AUs. The two directions are not transposes of each other.}
  \label{DISFA_DFER_heatmap}
\end{figure*}

\begin{figure*}[ht!]
  \centering
\includegraphics[width=\textwidth,height=\textheight,keepaspectratio]{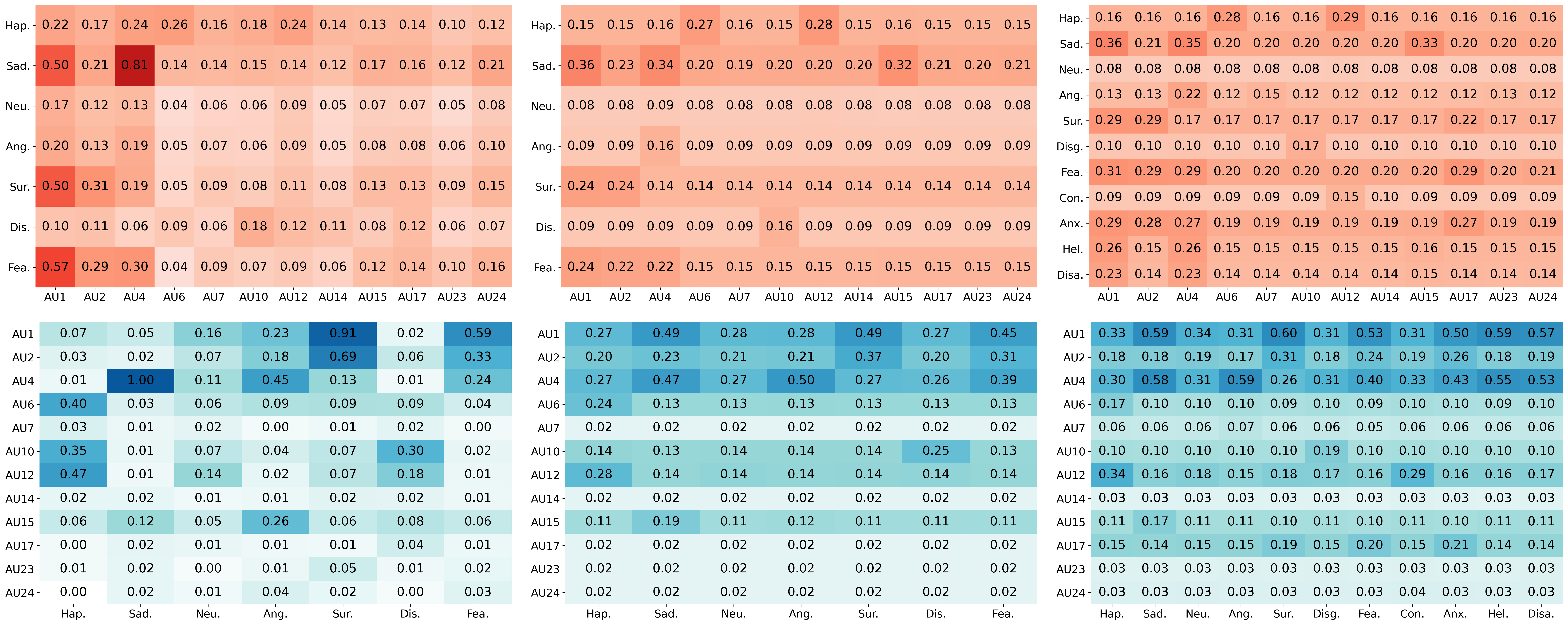}
\caption{The weight matrix activation maps, from left to right, represent joint learning of BP4D with DFEW, FERV39K, and MAFW, respectively. The first row shows the weights from AUs to expressions, and the second row shows the weights from expressions to AUs. The two directions are not transposes of each other.}
  \label{BP4D_DFER_heatmap}
\end{figure*}

Figs.~\ref{DISFA_DFER_heatmap} and \ref{BP4D_DFER_heatmap} visualize the bidirectional semantic mapping weights learned by DPM. The first row shows the contribution of AUs to expressions. The second row shows the contribution of expressions to AUs. The activation patterns in the two directions are not simple transposes of each other. This indicates that SSM learns directional and dynamic semantic mappings rather than static and symmetric prior correspondences. Meanwhile, several associations consistent with FACS priors remain stable across different dataset combinations. For example, happiness is associated with AU6 and AU12, surprise with AU1, AU2, and AU26, and disgust with AU9 and AU10. In contrast, the reverse-direction mappings exhibit stronger distributional characteristics and context dependence. This suggests that the constraints from expressions to AUs involve richer compositional structures. These visualizations qualitatively support the ability of DPM to preserve prior structure while achieving data-driven adaptation.

\subsection{Semantic Label Descriptions}
\label{supp-sec:Descriptions}

Table~\ref{tab:description} provides the textual construction basis of TSP. AU descriptions are directly adopted from the canonical FACS Manual descriptions and correspond to localized and atomic semantic units with explicit muscular-action meanings. Expression descriptions are compositionally constructed according to the prototypical AU--expression configurations summarized in the FACS Investigator's Guide while retaining only the AU labels available in the paired AU datasets (BP4D and DISFA). Therefore, these expression descriptions are dataset-constrained approximations rather than complete FACS prototypes. For non-basic MAFW categories not covered by the Guide, manually composed, dataset-constrained AU descriptions are used as initial semantic approximations. 

\begin{table*}[t]
\centering
\caption{Semantic label descriptions used in TSP for AU and expression categories. The upper section lists AU labels and their FACS-consistent atomic descriptions. The lower section presents expression labels, their AU combinations, and compositional semantic descriptions. The AU combinations are derived from the AUs available in BP4D and DISFA, and thus provide dataset-constrained approximations rather than full FACS prototypes.}
\label{tab:description}
\begin{adjustbox}{max width=\textwidth}
\small
\begin{tabular}{ccc}
\toprule
Label & \multicolumn{2}{c}{Description} \\
\midrule
AU1  & \multicolumn{2}{c}{inner brow raiser} \\
AU2  & \multicolumn{2}{c}{outer brow raiser} \\
AU4  & \multicolumn{2}{c}{brow lowerer} \\
AU6  & \multicolumn{2}{c}{cheek raiser} \\
AU7  & \multicolumn{2}{c}{lid tightener} \\
AU9  & \multicolumn{2}{c}{nose wrinkler} \\
AU10 & \multicolumn{2}{c}{upper lip raiser} \\
AU12 & \multicolumn{2}{c}{lip corner puller} \\
AU14 & \multicolumn{2}{c}{dimpler} \\
AU15 & \multicolumn{2}{c}{lip corner depressor} \\
AU17 & \multicolumn{2}{c}{chin raiser} \\
AU23 & \multicolumn{2}{c}{lip tightener} \\
AU24 & \multicolumn{2}{c}{lip presser} \\
AU25 & \multicolumn{2}{c}{lips part} \\
AU26 & \multicolumn{2}{c}{jaw drop} \\
\midrule
Label & AU Combination & Description \\
\midrule
Happiness      & AU6+AU12                       & cheek raiser, lip corner puller \\
Sadness        & AU1+AU4+AU15                   & inner brow raiser, brow lowerer, lip corner depressor \\
Neutral        & None                           & relaxed facial muscles, no significant action units \\
Anger          & AU4+AU7+AU23                   & brow lowerer, lid tightener, lip tightener \\
Surprise       & AU1+AU2+AU25+AU26                   & inner brow raiser, outer brow raiser, lips part, jaw drop \\
Disgust        & AU9+AU10+AU15                  & nose wrinkler, upper lip raiser, lip corner depressor \\
Fear           & AU1+AU2+AU4+AU7+AU26           & inner brow raiser, outer brow raiser, brow lowerer, lid tightener, jaw drop \\
Contempt       & AU12+AU14                      & lip corner puller, dimpler \\
Anxiety        & AU1+AU4+AU25                   & inner brow raiser, brow lowerer, lips part \\
Helplessness   & AU1+AU4+AU15+AU26              & inner brow raiser, brow lowerer, lip corner depressor, jaw drop \\
Disappointment & AU1+AU4+AU15+AU25              & inner brow raiser, brow lowerer, slight lip corner depressor, lips part \\
\bottomrule
\end{tabular}
\end{adjustbox}
\end{table*}

\begin{figure}[t]
    \centering
    \includegraphics[width=\columnwidth]{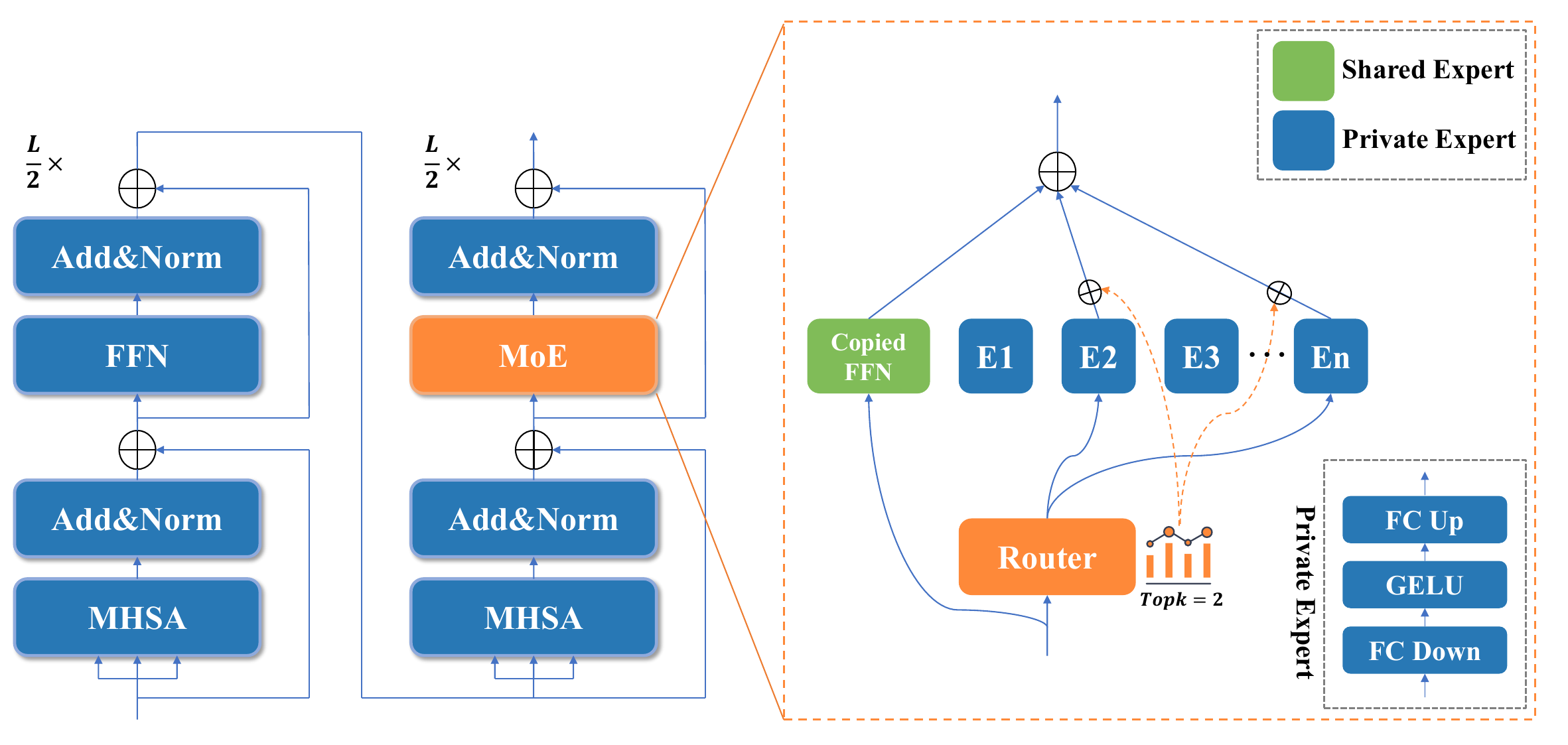}
    \caption{Proposed MoE in the CLIP visual encoder. Each Transformer block replaces the original FFN with an MoE layer containing one shared pretrained CLIP expert and multiple private experts.}
    \label{fig:supp_moe}
\end{figure}


\subsection{Mixture of Private Experts Module}
\label{supp-sec:Moe}
As illustrated in Fig.~\ref{fig:supp_moe}, for an input \(\boldsymbol{x}\in\mathbb{R}^d\) to a Transformer layer, we first apply layer normalization \(\widetilde{\boldsymbol{x}}=\mathrm{LN}(\boldsymbol{x})\). The router computes a score vector over \(m\) experts via a linear projection \(\boldsymbol{l} = R\,\widetilde{\boldsymbol{x}}\), and normalizes these scores with a softmax to obtain the gating vector \(\boldsymbol{g} = \mathrm{softmax}(\boldsymbol{l})\). To guarantee sparse computation and activate only a small subset of experts, we adopt a top-\(K\) selection strategy (in the figure \(K=2\)), and denote the selected expert set by \(\boldsymbol{S} = \operatorname{TopK}(\boldsymbol{g}, K)\). The \(j\)-th expert is implemented as a compact FFN:
\begin{equation}
\boldsymbol{y}_j = E_j(\widetilde{\boldsymbol{x}}) = \boldsymbol{W}_j^{(2)}\text{GELU}\bigl(\boldsymbol{W}_j^{(1)}\widetilde{\boldsymbol{x}} + \boldsymbol{b}_j^{(1)}\bigr) + \boldsymbol{b}_j^{(2)}.
\end{equation}

To stabilize training, the expert pool contains a shared expert \(E_s\) (initialized by copying CLIP's original FFN to preserve pretrained knowledge) and several private experts \(E_j\) (maintained separately for each task). The gating weights of the selected experts are re-normalized and their outputs are fused by a weighted sum:
\begin{equation}
\boldsymbol{y}= E_s(\widetilde{\boldsymbol{x}})+\gamma \sum_{j}\boldsymbol{y}_j,
\end{equation}
where $\gamma$ denotes a learnable vector.

\subsection{Sensitivity to the Task-Loss Weight}
\label{supp-sec:lambda-sensitivity}

Table~\ref{tab:lambda_sensitivity} studies the task-loss weight \(\lambda\), which controls the relative contribution of the AU loss in the joint objective. Increasing \(\lambda\) from \(0.25\) to \(2\) steadily improves AU detection. SSM achieves the best BP4D result and the best DFEW UAR/WAR under both dataset pairings when \(\lambda=2\). Although \(\lambda=4\) gives the highest DISFA F1 score, it reduces BP4D and DFEW performance. Therefore, \(\lambda=2\) provides the most balanced setting and is used by default.

\begin{table}[ht]
  \centering
  \scriptsize
  \setlength{\tabcolsep}{6pt} 
  \caption{Sensitivity to the task-loss weight \(\lambda\). The two DFEW columns report joint training with BP4D and DISFA, respectively. BP4D and DISFA: F1 score. DFEW: UAR/WAR.}
  \label{tab:lambda_sensitivity}
  \resizebox{\columnwidth}{!}{%
    \begin{tabular}{c|cc|cc}
      \toprule
      \(\lambda\) & \makecell{BP4D} & \makecell{DFEW} & \makecell{DISFA} & \makecell{DFEW}  \\
      \midrule
      0.25 & 67.5 &64.74/77.61 & 69.0 &65.50/77.83    \\
      0.50 & 67.6 &67.77/77.53 &69.8&65.73/77.57    \\
      1.0 & 67.8 &67.70/77.40 &70.3 &65.61/77.87   \\
      2.0 & \textbf{68.5} &\textbf{68.59}/\textbf{77.88} & 71.3 & \textbf{66.64}/\textbf{78.09} \\
      3.0 & 68.0 &68.24/77.48 & 71.2 & 66.62/77.40 \\
      4.0 & 67.8 &66.41/77.14 & \textbf{71.8} & 65.61/77.48 \\
      \bottomrule
    \end{tabular}%
  }
\end{table} 

\subsection{Effects of Auxiliary Data, FACS Knowledge, and Dynamic Adjustment}
\label{supp-sec:transfer-sources}

Tables~\ref{tab:fe_data_facs} and \ref{tab:au_data_facs} separate the effects of auxiliary-task data, fixed FACS-informed prior, and data-driven adjustment. Without auxiliary data, adding only the fixed prior produces negligible changes, with F1 score variations of -0.1\% and +0.1\% on BP4D and DISFA, respectively. Using all auxiliary data without the FACS-informed prior already improves BP4D (F1 score: +1.0\%) and DISFA (F1 score: +0.8\%), as well as DFEW (UAR: +1.27\%, WAR: +0.81\%). With the same FACS initialization, dynamic adjustment further improves BP4D (F1 score: +1.2\%) and DISFA (F1 score: +1.1\%) in FE$\rightarrow$AU transfer, and improves DFEW in AU$\rightarrow$FE transfer (UAR: +1.97\%, WAR: +0.83\%) when paired with BP4D. Moreover, using only 20\% auxiliary data with FACS initialization and dynamic adjustment already outperforms both the 100\% data-only setting and the 100\% data setting with fixed prior in both transfer directions. These results show that auxiliary-task data and fixed FACS-informed prior alone cannot fully explain the improvements, while data-driven adjustment further enables effective adaptation of the prior mapping under heterogeneous datasets.

\begin{table}[ht]
  \centering
  \scriptsize
  \setlength{\tabcolsep}{6pt} 
\caption{Comparison of auxiliary FE data, the FACS-informed prior knowledge, and dynamic updating for FE$\rightarrow$AU transfer. FE data are from DFEW.}
  \label{tab:fe_data_facs}
  \resizebox{\columnwidth}{!}{%
    \begin{tabular}{ccc|cc}
      \toprule
      \multicolumn{3}{c|}{Transfer Setting}
      & \multicolumn{2}{c}{AU Detection (F1)} \\
      \cmidrule(lr){1-3}\cmidrule(lr){4-5}
      FE Data & FACS Prior & Dynamic & BP4D & DISFA \\
      \midrule
      0\%   &          &          & 66.2 & 69.6 \\
      0\%   & \checkmark &          & 66.1 & 69.7 \\
      100\% &          &          & 67.2 & 70.4 \\
      100\% & \checkmark &          & 67.3 & 70.2 \\
      20\%  & \checkmark & \checkmark & 67.6 & 70.8 \\
      100\% & \checkmark & \checkmark &
      \textbf{68.5} & \textbf{71.3} \\
      \bottomrule
    \end{tabular}%
  }
\end{table}

\begin{table}[ht]
  \centering
  \scriptsize
  \setlength{\tabcolsep}{6pt} 
\caption{Comparison of auxiliary AU data, the FACS-informed prior knowledge, and dynamic updating for AU$\rightarrow$FE transfer. AU data are from BP4D.}
  \label{tab:au_data_facs}
  \resizebox{\columnwidth}{!}{%
    \begin{tabular}{ccc|cc}
      \toprule
      \multicolumn{3}{c|}{Transfer Setting}
      & \multicolumn{2}{c}{DFER on DFEW} \\
      \cmidrule(lr){1-3}\cmidrule(lr){4-5}
      AU Data & FACS Prior & Dynamic & UAR & WAR \\
      \midrule
      0\%   &          &          & 63.98 & 76.16 \\
      0\%   & \checkmark &         & 63.56 & 77.53 \\
      100\% &          &          & 65.25 & 76.97 \\
      100\% & \checkmark &          & 66.62 & 77.05 \\
      20\%  & \checkmark & \checkmark & 66.95 & 77.57 \\
      100\% & \checkmark & \checkmark &
      \textbf{68.59} & \textbf{77.88} \\
      \bottomrule
    \end{tabular}%
  }
\end{table}

\subsection{Failure Case Analysis}

To further analyze the limitations of SSM, we define a semantic-transfer failure as a DFER sample or an AU decision correctly predicted by Baseline but incorrectly predicted by SSM.

\begin{figure}[t]
    \centering
    \includegraphics[width=\columnwidth]{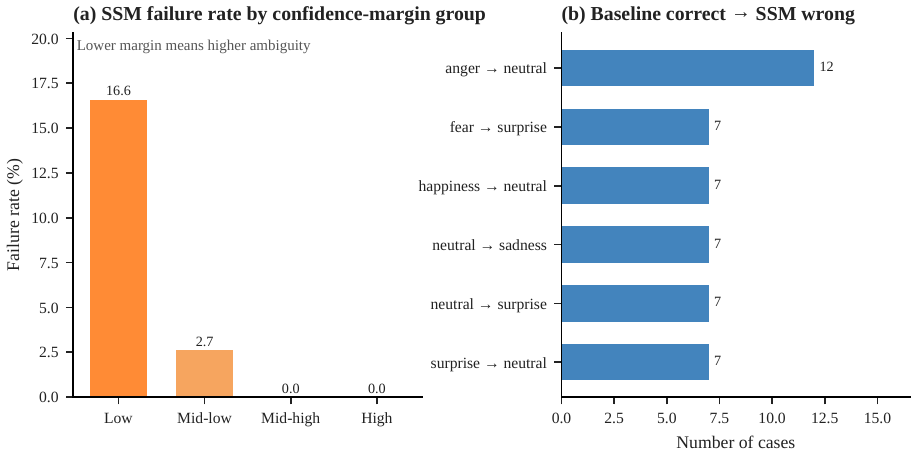}
    \caption{Failure analysis on DFEW fold 5. (a) SSM failure rate among Baseline-correct samples, computed within intervals of the Baseline confidence margin. The horizontal axis shows the confidence-margin intervals, and the vertical axis shows the proportion of samples misclassified by SSM in each interval. The confidence margin is the difference between Baseline’s top-1 and top-2 class probabilities; a smaller margin indicates greater ambiguity. (b) Most frequent Baseline-correct $\rightarrow$ SSM-wrong transitions.}
    \label{fig:dfer_failure_analysis}
\end{figure}

For the DFER task, Fig.~\ref{fig:dfer_failure_analysis}(a) shows that SSM failures are concentrated in ambiguous expressions. Among samples correctly predicted by Baseline, the failure rate decreases from 16.6\% in the Low-margin group to 2.7\%, 0.0\%, and 0.0\% as the confidence margin increases. This result indicates that semantic transfer is more likely to become misleading when the original prediction is ambiguous. Fig.~\ref{fig:dfer_failure_analysis}(b) shows the most frequent Baseline-correct $\rightarrow$ SSM-wrong transitions on DFEW fold 5. Five of the six most frequent transitions involve neutral as either the ground-truth or predicted class, indicating that prediction confusion with neutral is a common failure mode.

\begin{figure}[t]
    \centering
    \includegraphics[width=\columnwidth]{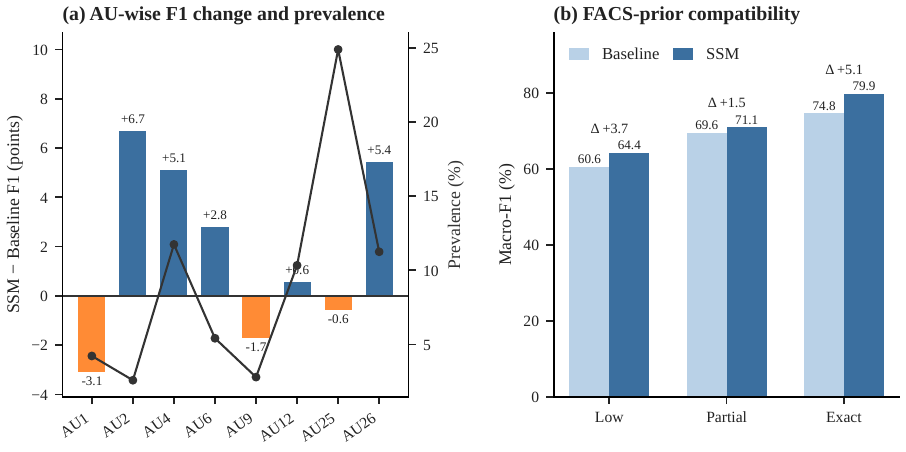}
    \caption{AU-level analysis on DISFA fold 1. (a) Bars show the AU-wise F1 change from Baseline to SSM (left vertical axis), and the black line shows test-set prevalence (right vertical axis). (b) Macro-F1 of Baseline and SSM at different levels of compatibility between the ground-truth active AU set and the FACS-informed AU combinations. Compatibility is the maximum Jaccard similarity to the six non-neutral expression prototypes; higher compatibility indicates less deviation from the FACS-informed prior. Low, Partial, and Exact denote Jaccard similarities $<0.5$, $[0.5,1.0)$, and $1.0$, respectively.}
    \label{fig:au_failure_analysis}
\end{figure}

For the AU detection task, Fig.~\ref{fig:au_failure_analysis}(a) shows non-uniform transfer across AUs. The F1 scores increase for five of the eight AUs, with the largest gains on AU2 (+6.7 points) and AU26 (+5.4 points), whereas the scores decrease for AU1 (-3.1 points), AU9 (-1.7 points), and AU25 (-0.6 points). These changes do not follow a clear prevalence-related pattern, indicating that rarer AUs do not necessarily suffer greater negative transfer. In Fig.~\ref{fig:au_failure_analysis}(b), frames are grouped according to the maximum Jaccard similarity between their ground-truth active AU sets and the AU combinations of the non-neutral expression prototypes. Compared with Baseline, SSM improves Macro-F1 by 3.7, 1.5, and 5.1 points in the Low, Partial, and Exact groups, respectively. The smaller gains in the Low and Partial groups suggest that deviations from the FACS prior weaken the transfer benefit.

\begin{figure*}[t]
    \centering
    \includegraphics[width=\textwidth]{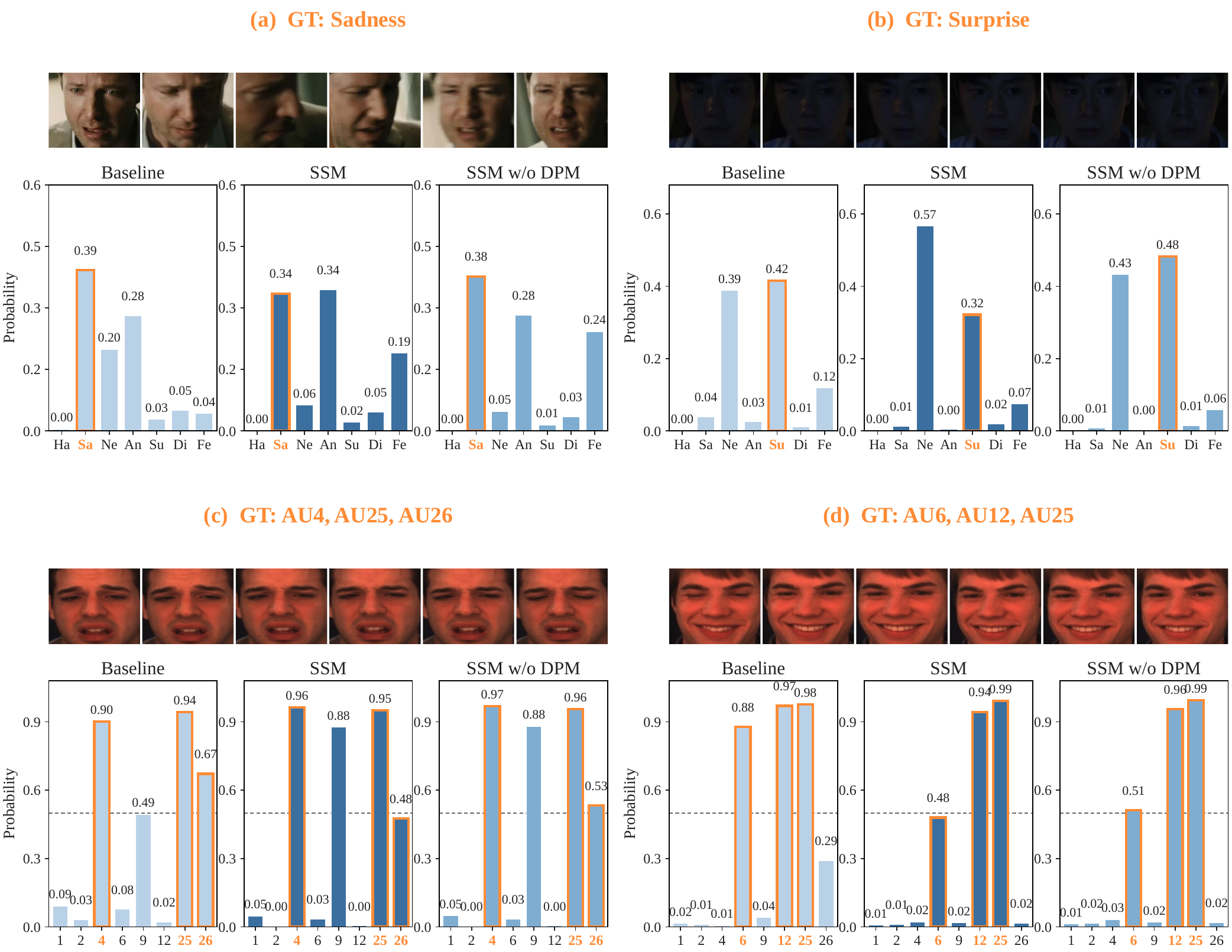}
    \caption{Examples correctly predicted by Baseline but incorrectly predicted by SSM (Set 1). The upper row shows two independent DFEW cases, and the lower row shows two independent DISFA cases. Orange outlines mark the ground-truth labels. In the AU panels, the dashed line marks the decision threshold of 0.5. ``SSM w/o DPM'' uses the same trained SSM checkpoint with DPM disabled at inference.}
    \label{fig:semantic_transfer_failures_1}
\end{figure*}

\begin{figure*}[t]
    \centering
    \includegraphics[width=\textwidth]{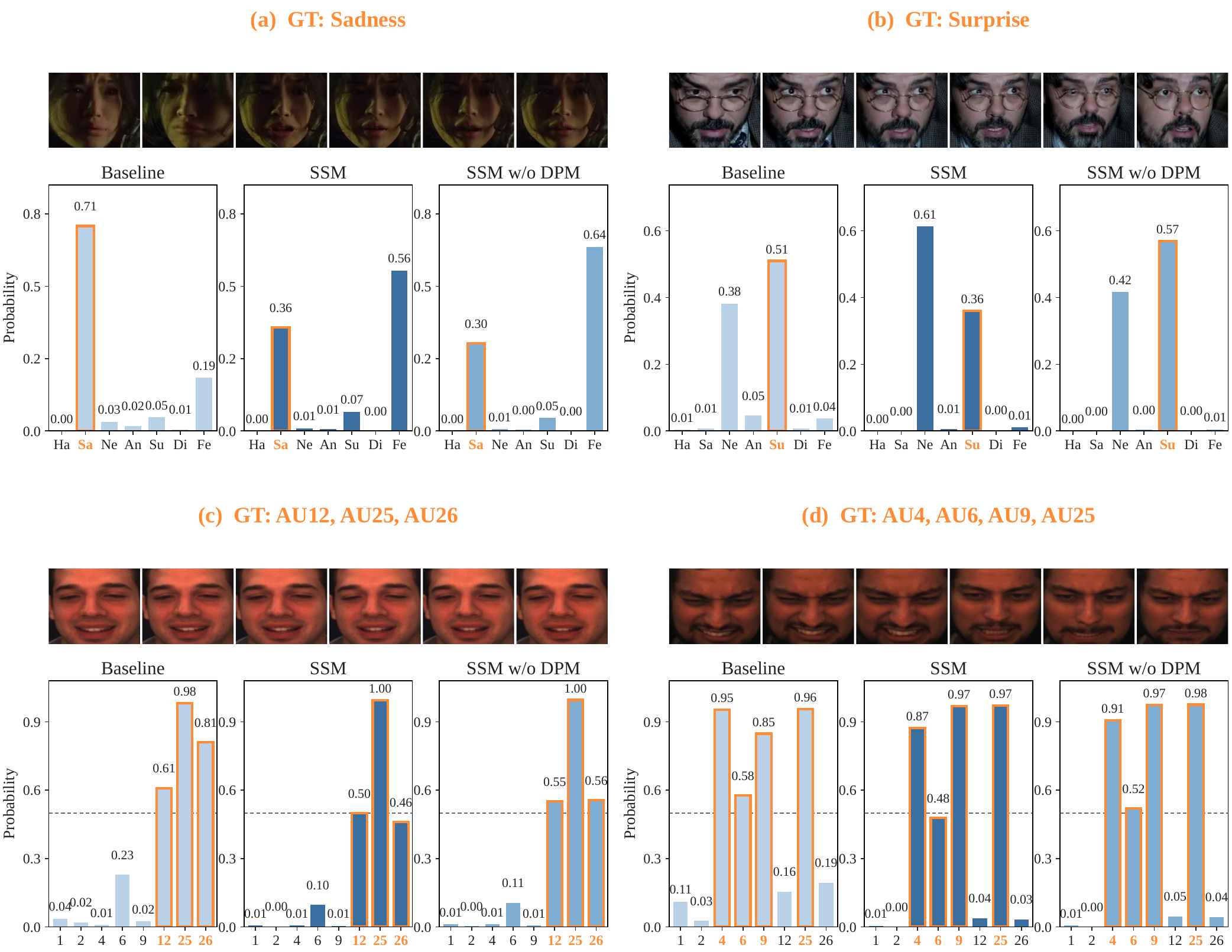}
    \caption{Additional examples correctly predicted by Baseline but incorrectly predicted by SSM (Set 2). The upper row shows two independent DFEW cases, and the lower row shows two independent DISFA cases. Orange outlines mark the ground-truth labels. In the AU panels, the dashed line marks the decision threshold of 0.5. ``SSM w/o DPM'' uses the same trained SSM checkpoint with DPM disabled at inference.}
    \label{fig:semantic_transfer_failures_2}
\end{figure*}

Figs.~\ref{fig:semantic_transfer_failures_1} and \ref{fig:semantic_transfer_failures_2} present eight failure cases, including four DFER cases and four AU cases. Using the same trained SSM checkpoint, we disable DPM only at inference. This restores the correct prediction in seven cases; only the sadness$\rightarrow$fear case remains incorrect. The DFER failures are concentrated among ambiguous expressions, while the AU detection failures include ground-truth AU combinations that deviate from the FACS-informed prior. These results suggest that the current dataset-level mapping may be less reliable for ambiguous expressions and atypical AU activation patterns.

\subsection{Computational Complexity}
\label{supp-sec:complexity}

Table~\ref{tab:complexity} reports the computational complexity under a controlled heterogeneous mini-batch forward setting. Compared with Baseline, SSM adds \(0.008\)M trainable parameters and \(0.73\%\) FLOPs, and increases inference time by \(1.56\%\). The increase in total parameters mainly comes from the frozen CLIP text encoder. Baseline + TSP and full SSM have the same parameter and FLOP counts at the reported precision, indicating that DPM introduces negligible additional parameters and FLOPs.

\begin{table*}[t]
\centering
\caption{Computational complexity under the heterogeneous mini-batch setting. DFER and AU batches contain 12 and 8 clips, respectively, with 16 frames per clip; thus joint models process 320 video frames per forward pass. ``Frames'' denotes the number of video frames processed in one forward pass. ``Total'' and ``Train'' denote total and trainable parameters. ``Time'' denotes the average inference time over 50 forward passes after 10 warm-up runs under the \texttt{torch.no\_grad()} context manager.}
\label{tab:complexity}

\renewcommand{\arraystretch}{1.15}
\setlength{\tabcolsep}{8pt}

\begin{tabular*}{\textwidth}{@{\extracolsep{\fill}}lccccc}
\hline
Model & Frames & Total(M) & Train(M) & FLOPs(G) & Time(ms) \\
\hline
STL-DFER     & 192 & 88.308 & 88.308 & 3238.905 & 253.251 \\
STL-AU       & 128 & 88.308 & 88.308 & 2159.253 & 169.090 \\
Baseline     & 320 & 109.346 & 109.346 & 5994.240 & 487.113 \\
Baseline + TSP & 320 & 147.485 & 109.354 & 6037.918 & 494.001 \\
Baseline + TSP + DPM (SSM)         & 320 & 147.485 & 109.354 & 6037.918 & 494.721 \\
\hline
\end{tabular*}
\end{table*}

\begin{table}[t]
\centering
\caption{Quantitative comparison of cross-dataset representation discrepancy among STL, Baseline, and SSM. \(\mathrm{MMD}^{2}\) and Cross/Intra are computed based on the 512-dimensional encoder embeddings before the classification heads. An equal number of DFEW and DISFA samples are used, with the same sampled instances across all three settings. Lower values indicate a smaller representation discrepancy.}
\label{bg_mmd}
\scriptsize
\setlength{\tabcolsep}{8pt}
\begin{tabular}{ccc}
\toprule
Setting & \(\mathrm{MMD}^{2}\) & Cross/Intra \\
\midrule
STL      & 0.2656 & 1.3046 \\
Baseline & 0.1764 & 1.1769 \\
SSM      & 0.1459 & 1.1374 \\
\bottomrule
\end{tabular}
\end{table}

\subsection{Heterogeneous-Dataset Representation Analysis}
\label{supp-sec:domain-gap}

\begin{figure*}[!t]
  \centering
  \includegraphics[width=\textwidth]{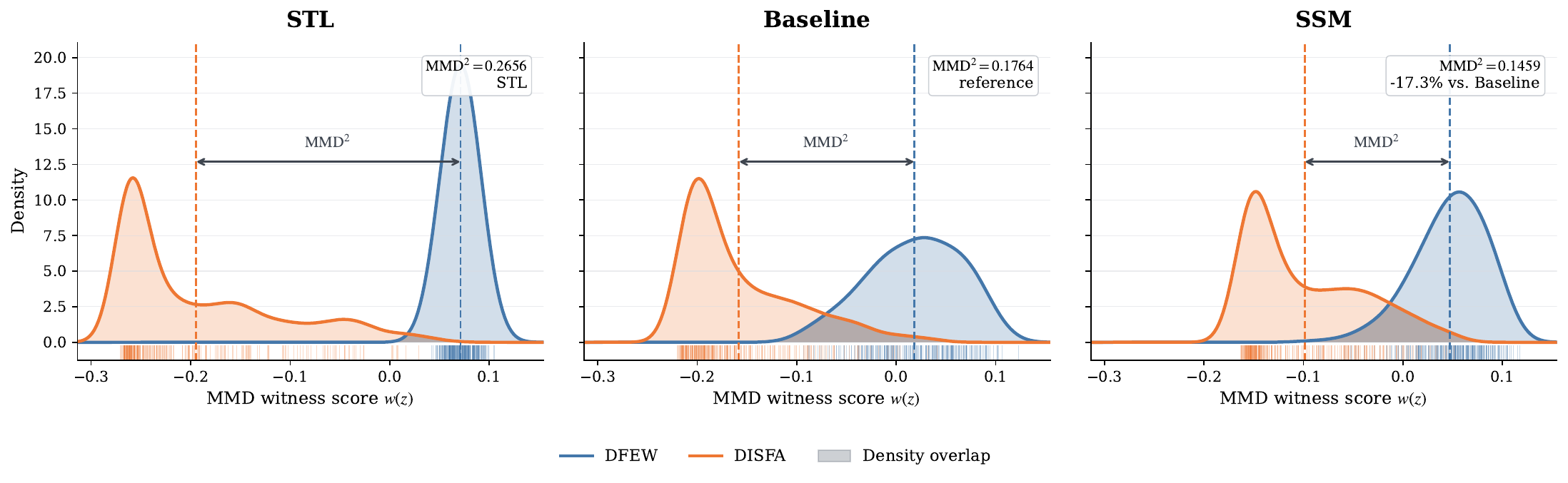}
  \caption{Visualization of the DFEW--DISFA representation gap under different training settings. The density curves show the distributions of maximum mean discrepancy (MMD) witness scores computed with an RBF kernel from the 512-dimensional encoder embeddings. A witness score reflects an embedding's relative similarity between the DFEW and DISFA distributions. Dashed lines indicate the domain-wise means. Compared with STL (\(0.2656\)) and the joint Baseline (\(0.1764\)), SSM reduces the value of \(\mathrm{MMD}^{2}\) to \(0.1459\), corresponding to a \(17.3\%\) relative reduction from Baseline and indicating a smaller cross-domain representation discrepancy.}
  \label{fig:domain-gap}
\end{figure*}

Fig.~\ref{fig:domain-gap} visualizes the representation discrepancy between DFEW and DISFA using the distributions of their RBF-MMD witness scores. STL employs independently trained task-specific encoders, whereas Baseline and SSM use jointly trained visual encoders. The dashed lines denote the domain-wise mean witness scores, whose separation equals the unbiased empirical \(\mathrm{MMD}^{2}\). Compared with STL (\(\mathrm{MMD}^{2}=0.2656\)), Baseline reduces \(\mathrm{MMD}^{2}\) to \(0.1764\), indicating that shared visual learning already reduces the measured representation discrepancy. SSM further decreases \(\mathrm{MMD}^{2}\) to \(0.1459\), corresponding to a \(17.3\%\) relative reduction compared with Baseline. These results indicate that SSM further mitigates the measured representation discrepancy between the heterogeneous datasets.

Table~\ref{bg_mmd} reports the representation discrepancy between randomly and equally sampled DFEW and DISFA samples in STL, Baseline, and SSM measured by \(\mathrm{MMD}^{2}\) \cite{gretton2012kernel} and Cross/Intra \cite{mcclain1975clustisz}. Both measures are computed based on the 512-dimensional encoder embeddings before the classification heads. STL uses independently trained task-specific encoders, whereas Baseline and SSM use jointly trained visual encoders. Compared with STL, Baseline reduces \(\mathrm{MMD}^{2}\) from \(0.2656\) to \(0.1764\) and Cross/Intra from \(1.3046\) to \(1.1769\). SSM further reduces the two measures to \(0.1459\) and \(1.1374\), respectively. These results indicate that SSM further mitigates the representation discrepancy between the heterogeneous datasets.

\begin{figure}[!t]
  \centering
  \includegraphics[width=\columnwidth]{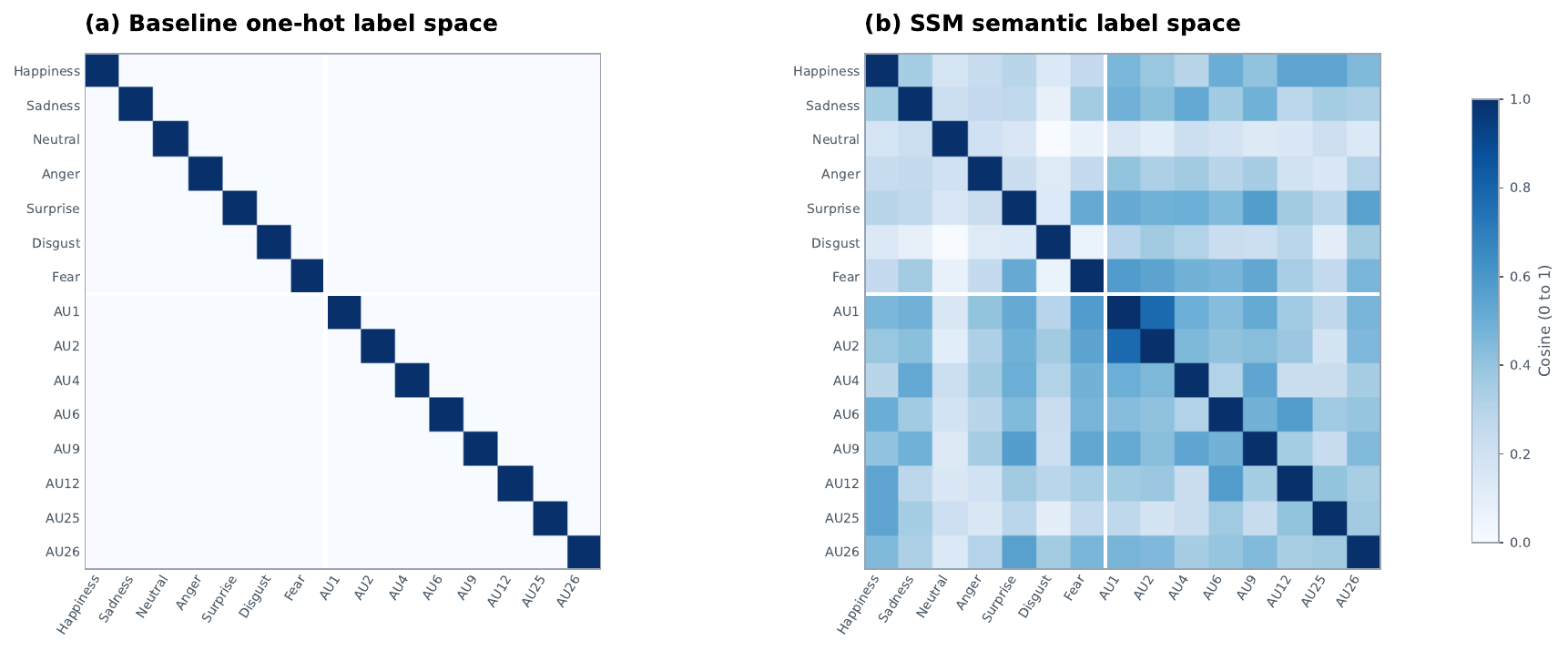}
  \caption{Comparison of the AU and FE label spaces in Baseline (left) and SSM (right), using the DFEW and DISFA datasets. A matrix entry gives the cosine similarity between two label representations. Baseline uses independent one-hot codes, whereas SSM uses the textual semantic codes.}
  \label{fig:label-space}
\end{figure}

Besides, Fig.~\ref{fig:label-space} compares the AU and FE label spaces. In Baseline, the off-diagonal cosine similarities between the independent one-hot AU and FE labels are zero. In SSM, the off-diagonal cosine similarities between the final normalized AU and FE textual prototypes reflect varying degrees of inherent semantic relatedness among different AU and FE label pairs. Thus, SSM also mitigates the gap between the AU and FE label spaces, in addition to reducing the measured discrepancy between their visual representations.

\FloatBarrier
\putbib[main]
\end{bibunit}

\end{document}